\documentclass[11pt,a4paper]{scrartcl}
\usepackage[round]{natbib}

\usepackage[top=3cm, bottom=3cm, left=3cm, right=3cm]{geometry}

\usepackage{booktabs}

\usepackage{color}
\usepackage{latexsym}              
\usepackage{amsmath}               
\usepackage{amssymb}               
\usepackage{amsfonts}              
\usepackage{amsthm}                
\usepackage{multirow}
\usepackage{tikz}                  
\usetikzlibrary{arrows,positioning,shapes}
\usetikzlibrary{bayesnet}
\usepackage{tcolorbox}
\usepackage{dsfont}
\usepackage{upgreek}
\usepackage{graphicx}
\usepackage{xspace} 
\usepackage{float}
\usepackage{cuted} 
\usepackage{algorithm, algorithmic}
\restylefloat{table}

\usepackage[english]{babel} 

\RequirePackage[%
  pdfstartview=FitH,%
  breaklinks=true,%
  bookmarks=true,%
  colorlinks=true,%
  linkcolor= blue,
  anchorcolor=blue,%
  citecolor=blue,
  filecolor=blue,%
  menucolor=blue,%
  urlcolor=blue%
  ]{hyperref}
  
\newcommand{\paren}[1]{\left( #1 \right)}
\newcommand{\croch}[1]{\left[\, #1 \,\right]}
\newcommand{\acc}[1]{\left\{ #1 \right\}}
\newcommand{\abs}[1]{\left| #1 \right|}

\newcommand{\argmax}{\operatornamewithlimits{argmax}} 
\newcommand{\dif}{\mathop{}\!\mathrm{d}}
\renewcommand\hat{\widehat}

\newcommand{\N}{\mathcal{N}}
\renewcommand{\vert}{\mid}
\newcommand{\tr}[1]{\textrm{tr}\left(#1\right)}

\newcommand{\I}{\mathcal{I}}
\newcommand{\T}{\mathcal{T}}
\newcommand{\K}{\mathcal{K}}

\newcommand{\algo}{\textsc{Magma}\xspace}
\newcommand{\algoclust}{\textsc{MagmaClust}\xspace}

\newcommand\Hki{\mathcal{H}_{k}^{i}}

\newcommand\yi{\mathbf{y}_i}
\newcommand\yst{y_{*}}
\newcommand\yii{\acc{\mathbf{y}_i}_i}

\newcommand\ti{\mathbf{t}_i}
\newcommand\Ut{\mathbf{t}}
\newcommand\tpred{\mathbf{t}^{p}}
\newcommand\tst{\mathbf{t}_{*}}
\newcommand\tpst{\mathbf{t}^{p}_{*}}
\newcommand\mub{ \boldsymbol{\mu} }
\newcommand\muk{ \mu_k }
\newcommand\mukt{ \mu_k(\mathbf{t}) }
\newcommand\mukti{ \mu_k(\mathbf{t}_i) }
\newcommand\mukk{\acc{ \mu_k(\cdot)}_k }
\newcommand\mukkt{\acc{ \mu_k(\mathbf{t})}_k }
\newcommand\mukkti{\acc{ \mu_k(\mathbf{t}_i)}_k }

\newcommand\mukktpst{\acc{ \mu_k(\mathbf{t}_{*}^p)}_k }

\newcommand\mkhat{\hat{m}_k}

\newcommand\Z{\mathbf{Z}}
\newcommand\Zik{Z_{ik}}
\newcommand\Zi{\mathbf{Z}_i}
\newcommand\Zii{\acc{\mathbf{Z}_i}_i}
\newcommand\Zst{\mathbf{Z}_*}
\newcommand\Zstk{Z_{*k}}

\newcommand\mk{m_k}
\newcommand\mkt{m_k(\mathbf{t})}

\newcommand\psii{\psi_{\theta_i, \sigma_i^2}}
\newcommand\psiihat{\psi_{\hat{\theta}_i, \hat{\sigma}_i^2}}
\newcommand\Psii{\boldsymbol{\Psi}_{\theta_i, \sigma_i^2}}
\newcommand\Psiiti{{\boldsymbol{\Psi}_{\theta_i, \sigma_i^2}^{\mathbf{t}_i}}}
\newcommand\Psiihat{\boldsymbol{\Psi}_{\hat{\theta}_i, \hat{\sigma}_i^2}}
\newcommand\Psiihatti{{\boldsymbol{\Psi}_{\hat{\theta}_i, \hat{\sigma}_i^2}^{\ti}}}
\newcommand\Psitilde{{\tilde{\boldsymbol{\Psi}}_i}}

\newcommand\thetaii{\acc{\theta_i}_i}
\newcommand\thetaiihat{\acc{\hat{\theta}_i}_i}
\newcommand\sigmaii{\acc{\sigma_i^2}_i}
\newcommand\sigmaiihat{\acc{\hat{\sigma}_i^2}_i}

\newcommand\Ckt{{\mathbf{C}_{\gamma_k}^{\mathbf{t}}}}

\newcommand\Ckhatt{ {\mathbf{C}_{\hat{\gamma}_k}^{\mathbf{t}}} }

\newcommand\Chatk{\mathbf{\hat{C}}_k}
\newcommand\Chatkt{\mathbf{\hat{C}}_k^{\mathbf{t}}}

\newcommand\Chatkti{\mathbf{\hat{C}}_k^{\mathbf{t}_i}} 
\newcommand\Chatktst{\mathbf{\hat{C}}_k^{\mathbf{t}_*}} 
\newcommand\Chatlti{\mathbf{\hat{C}}_l^{\mathbf{t}_i}} 
 
\newcommand\Chatltst{\mathbf{\hat{C}}_l^{\mathbf{t}_*}}
 
\newcommand\Ckhattpst{ {\mathbf{C}_{\hat{\gamma}_k}^{\tpst}} }
\newcommand\Chatktpst{\mathbf{\hat{C}}_k^{\mathbf{t}_{*}^p}} 

\newcommand\sigmai{\sigma_i^2}
\newcommand\sigmaihat{\hat{\sigma}_i^2}
\newcommand\thetai{\theta_i}
\newcommand\thetaihat{\hat{\theta}_i}
\newcommand\sigmast{\sigma_*^2}
\newcommand\sigmasthat{\hat{\sigma}_*^2}
\newcommand\thetast{\theta_*}
\newcommand\thetasthat{\hat{\theta}_*}
\newcommand\gammak{\gamma_k}
\newcommand\gammakhat{\hat{\gamma}_k}
\newcommand\gammakk{\acc{\gamma_k}_k}
\newcommand\gammakkhat{\acc{\hat{\gamma}_k}_k}
\newcommand\pii{\boldsymbol{\pi}}
\newcommand\pik{\pi_{k}}
\newcommand\piihat{\hat{\boldsymbol{\pi}}}
\newcommand\pikhat{\hat{\pi}_{k}}
\newcommand\pilhat{\hat{\pi}_{l}}
\newcommand\taui{\boldsymbol{\uptau}_{i}}
\newcommand\taust{\boldsymbol{\uptau}_{*}}
\newcommand\tauik{\uptau_{ik}}
\newcommand\taustk{\uptau_{*k}}

\newcommand\sumi{\sum\limits_{i = 1}^{M}}
\newcommand\sumk{\sum\limits_{k = 1}^{K}}
\newcommand\prodi{\prod\limits_{i = 1}^{M}}
\newcommand\prodk{\prod\limits_{k = 1}^{K}}

\newcommand\Hoo{\mathcal{H}_{00}}
\newcommand\Hoi{\mathcal{H}_{0i}}
\newcommand\Hko{\mathcal{H}_{k0}}
\renewcommand\Hki{\mathcal{H}_{ki}}  
  
  \AtBeginDocument{%
  \hypersetup{%
    pdfauthor={Leroy, Latouche, Guedj and Gey},%
    	colorlinks = true,%
  	urlcolor = blue,%
  	linkcolor = orange,%
  	citecolor = orange,%
    pdftitle={Cluster-Specific Predictions with Multi-Task Gaussian Processes - compilation: \today}%
  }
}

\usepackage{cleveref}
\crefname{assumption}{Assumption}{Assumptions}
\crefname{equation}{Eq.}{Eqs.}
\crefname{figure}{Fig.}{Figs.}
\crefname{table}{Table}{Tables}
\crefname{section}{Sec.}{Secs.}
\crefname{theorem}{Thm.}{Thms.}
\crefname{lemma}{Lemma}{Lemmas}
\crefname{corollary}{Cor.}{Cors.}
\crefname{example}{Example}{Examples}
\crefname{appendix}{Appendix}{Appendixes}
\crefname{remark}{Remark}{Remark}

\renewenvironment{proof}[1][\proofname]{{\bfseries #1.}}{\qed \\ }

\makeatother

\theoremstyle{plain}  
\newtheorem{theorem}{Theorem}[section]

\newtheorem{lemma}[theorem]{Lemma}
\newtheorem{proposition}[theorem]{Proposition}

\allowdisplaybreaks

\begin{document}
\title{Cluster-Specific Predictions with Multi-Task Gaussian Processes}

\author{\textbf{Arthur Leroy} \\ [2ex]
Department of Computer Science, The University of Manchester, \\
 United Kingdom\\
              \texttt{arthur.leroy.pro@gmail.com} \\\\
\textbf{Pierre Latouche} \\ [2ex]
Université de Paris, CNRS, MAP5 UMR 8145, \\
              F-75006 Paris, France \\
              \texttt{pierre.latouche@u-paris.fr} \\\\
\textbf{Benjamin Guedj} \\ [2ex]
              Inria, France and \\
               University College London, United Kingdom \\
              \texttt{benjamin.guedj@inria.fr} \\\\
\textbf{Servane Gey} \\ [2ex]
              Université de Paris, CNRS, MAP5 UMR 8145, \\
              F-75006 Paris, France \\
              \texttt{servane.gey@u-paris.fr} \\\\
}
\date{\today}

\maketitle

\begin{abstract}
A model involving Gaussian processes (GPs) is introduced to simultaneously handle multi-task learning, clustering, and prediction for multiple functional data. This procedure acts as a model-based clustering method for functional data as well as a learning step for subsequent predictions for new tasks.
The model is instantiated as a mixture of multi-task GPs with common mean processes. 
A variational EM algorithm is derived for dealing with the optimisation of the hyper-parameters along with the hyper-posteriors' estimation of latent variables and processes.  
We establish explicit formulas for integrating the mean processes and the latent clustering variables within a predictive distribution, accounting for uncertainty on both aspects. 
This distribution is defined as a mixture of cluster-specific GP predictions, which enhances the performances when dealing with group-structured data.
The model handles irregular grids of observations and offers different hypotheses on the covariance structure for sharing additional information across tasks.
The performances on both clustering and prediction tasks are assessed through various simulated scenarios and real data sets.
The overall algorithm, called \algoclust, is publicly available as an R package. 
\end{abstract}


\section{Introduction}
\label{sec:intro}


The classic context of regression aims at inferring the underlying mapping function associating input to output data.
In a probabilistic framework, some strategies assume that this function is drawn from a prior Gaussian process (GP).
According to \cite{RasmussenGaussianprocessesmachine2006a}, a Gaussian process can be defined as a collection of random variables (indexed over a continuum), any finite number of which having a joint Gaussian distribution.
From this definition, we may enforce some properties for the target function solely by characterising the mean and covariance parameters of the process. 
Since GPs are an example of kernel methods, a broad range of assumptions can be expressed through the definition of the covariance function.
We refer to \cite{DuvenaudAutomaticmodelconstruction2014} for a comprehensive review.
Despite undeniable advantages, natural implementations for GPs scale cubically with the number of data points, which constitutes a major drawback in many applications.
Thereby, the early literature focused on deriving tractable approximations to mitigate this problem \citep{SchwaighoferLearningGaussianProcess2004,
SnelsonSparseGaussianProcesses2006,TitsiasVariationalLearningInducing2009,HensmanGaussianProcessesBig2013}.
Subsequent reviews \citep{Quinonero-CandelaApproximationMethodsGaussian2007,BauerUnderstandingProbabilisticSparse2016} have also provided standardised formulations and comparisons on this topic.
Besides, several approximations have been proposed \citep{NealMonteCarloImplementation1997} and implemented \citep{RasmussenGaussianprocessesmachine2010,VanhataloGPstuffBayesianModeling2013} for tackling the issue of non-Gaussian errors and adapting GPs to a broad variety of likelihoods.
Since a GP corresponds to a probability distribution over a functional space, alternate approaches for modelling functional data \citep{RamsayFunctionalDataAnalysis2005} should also be mentioned, in particular for our clustering purpose. 
\newline

Unsupervised learning of functional data, also called \emph{curve clustering}, focuses on the definition of sub-groups of curves, making sense according to an appropriate measure of similarity. 
When dealing with functional data, the concept of basis functions expansion is of paramount importance for defining smooth and manageable representations of the data. 
Such a representation allows the adaptation of multivariate methods such as k-means, in combination with B-splines bases for instance  \citep{AbrahamUnsupervisedCurveClustering2003}, to handle curve clustering problems.
Different bases can be used, such as Fourier or wavelets \citep{GiacofciWaveletBasedClusteringMixedEffects2013}, according to the context and the nature of the signal.
Besides, model-based clustering methods aims at defining probabilistic techniques for this task, and many approaches \citep{JiangClusteringRandomCurves2012, JacquesFunclustcurvesclustering2013a} have been proposed in this sense for the past decade.
In particular, the algorithms \emph{funHDDC} \citep{Bouveyron2011} and \emph{funFEM} \citep{bouveyron2015discriminative} establish a mixture model where representative coefficients of the curves are supposed to come from cluster-specific Gaussian distributions.
Furthermore, the authors in \cite{SchmutzClusteringmultivariatefunctional2018} introduced an extension to the case of multivariate functional data.
A comprehensive review \citep{JacquesFunctionaldataclustering2014} has been proposed to discuss and compare the major approaches of this active research area. 
We can also mention recent works leveraging generalised Bayesian predictors and PAC-Bayes for learning and clustering streams of data \citep{LiquasiBayesianperspectiveonline2018,GuedjSequentialLearningPrincipal2019}, which later inspired a work on clustering \citep{pmlr-v130-cohen-addad21a}.
In line with the previous methods that simultaneously analyse multiple curves, we also introduce a framework that takes advantage of similarities between resembling data.
\newline 

The \emph{multi-task} paradigm consists in using data from several \emph{tasks} (also called \emph{batches} or \emph{individuals}) to improve the learning or predictive capacities compared to an isolated model. 
It was introduced by \cite{CaruanaMultitaskLearning1997} and then adapted in many fields of machine learning. 
An initial GP adaptation \citep{SchwaighoferLearningGaussianProcess2004} came as a hierarchical Bayesian model using an expectation-maximisation (EM) algorithm for learning, and a similar approach can be found in \cite{ShiHierarchicalGaussianprocess2005}.
Another hierarchical formulation using a GP to model the mean parameter of another GP was later proposed in \cite{hensman2013hierarchical}. 
Such modelling assumptions resemble those of the present paper, although the strategies used for learning and prediction largely differ.
Meanwhile, \cite{YuLearningGaussianProcesses2005} offered an extensive study of the relationships between the linear model and GPs to develop a multi-task formulation. 
More recently, the expression \emph{multi-task GP} has been coined by \cite{BonillaMultitaskGaussianProcess2008} for referring to a covariance structure involving inputs and tasks in two separate matrices.
Some further developments on this approach were proposed \citep{HayashiSelfmeasuringSimilarityMultitask2012,RakitschItallnoise2013,ChenGeneralizedConvolutionSpectral2020} and we can also mention the work of \cite{SwerskyMultiTaskBayesianOptimization2013} on Bayesian hyper-parameter optimisation in such models. 
Generally, as presented in the review \cite{AlvarezKernelsVectorValuedFunctions2012a} which favours the term \emph{multi-output GP}, all these frameworks can be expressed as specific cases of the \emph{linear model of coregionalisation} (LMC) introduced by \cite{goovaerts1997geostatistics} in the field of geostatistics.
Finally, let us emphasise the algorithm \algo from \cite{LeroyMAGMAInferencePrediction2020} that recently proposed a different multi-task paradigm for GPs, by transferring information through a latent mean process rather than the covariance structure, an intuition that partially appeared before in \cite{ShiGaussianProcessFunctional2007}. 
This approach offers enhanced performances in forecasting while scaling linearly in the number of tasks, which is noticeably lower than the previous multi-output methods which generally bear a cubic complexity. 
However, the assumption of a unique mean process might happen to be too restrictive and could benefit from a more general formulation. For instance, \cite{ShiCurvepredictionclustering2008} proposed an idea close to our following model by introducing a curve clustering component to a hybrid splines-GPs multi-task framework, although their approach does not fully account for uncertainty and cannot handle irregular measurements. Moreover, no implementation has been released for their algorithm, and by deriving a unified multi-task GP framework that is more general, we aim at offering to practitioners a powerful tool for tackling the simultaneous clustering and prediction of multiple functional data. 
\newline

\paragraph{Our contributions.} The present paper contributes a significant extension of \algo \citep{LeroyMAGMAInferencePrediction2020}, by introducing a clustering component into the procedure.
To this end, \textbf{(i)} we introduce a more general model involving multiple mean GPs, each one being associated with a particular cluster. 
These processes represent the prior mean trend, possibly different from one cluster to another, that is associated with an individual covariance structure for each functional data.
Moreover, we propose 4 different modelling hypotheses regarding the kernels' hyper-parameters of the GPs.
\textbf{(ii)} We  derive a variational expectation-maximisation (VEM)  algorithm called \algoclust
(available as an R package on the CRAN and at \url{https://github.com/ArthurLeroy/MagmaClustR})  for  estimating  the hyper-parameters along  with the hyper-posterior distributions  of the
mean  processes and  latent  clustering variables.  A variational BIC criterion is proposed to estimate the number of clusters. 
\textbf{(iii)} We enrich this learning step with an additional EM algorithm and analytical formulas to determine both  clusters probabilities and predictive distributions for any new, partially observed, individual.
The final multi-task prediction can be expressed in terms of cluster-specific distributions or as an overall GPs mixture. 
The algorithmic complexity of learning and prediction steps are discussed as well. 
\textbf{(iv)} We illustrate the advantages of our approach on synthetic and three real-life data sets. 
The results exhibit that \algoclust outperforms state-of-the-art alternatives on both curve clustering and prediction tasks, in particular for group-structured data sets. 

\paragraph{Outline of the paper.} 
We introduce the multi-task Gaussian processes mixture model in \Cref{sec:modelling}, along with notation. \Cref{sec:inference} is devoted to the inference procedure, with a Variational Expectation-Maximisation (VEM) algorithm to estimate hyper-parameters and approximation of hyper-posterior distributions along with mixture proportions.
We leverage this strategy in \Cref{sec:prediction} and derive both a mixture and cluster-specific GP prediction formulas, for which we provide an analysis along with computational costs in \Cref{sec:complexity}.
The performances of our algorithm for clustering and prediction purposes are illustrated in \Cref{sec:exp} with a series of experiments on both synthetic and real-life data sets along with comparisons to competing state-of-the-art algorithms.
We close with a summary of our work in \Cref{sec:conclusion}. 
All proofs to original results are deferred to \Cref{sec:proofs}.

\section{Modelling}
\label{sec:modelling}

\subsection{Motivation}
\label{sec:motivation}
\begin{figure*}
    \begin{center}
       \includegraphics[width=\textwidth]{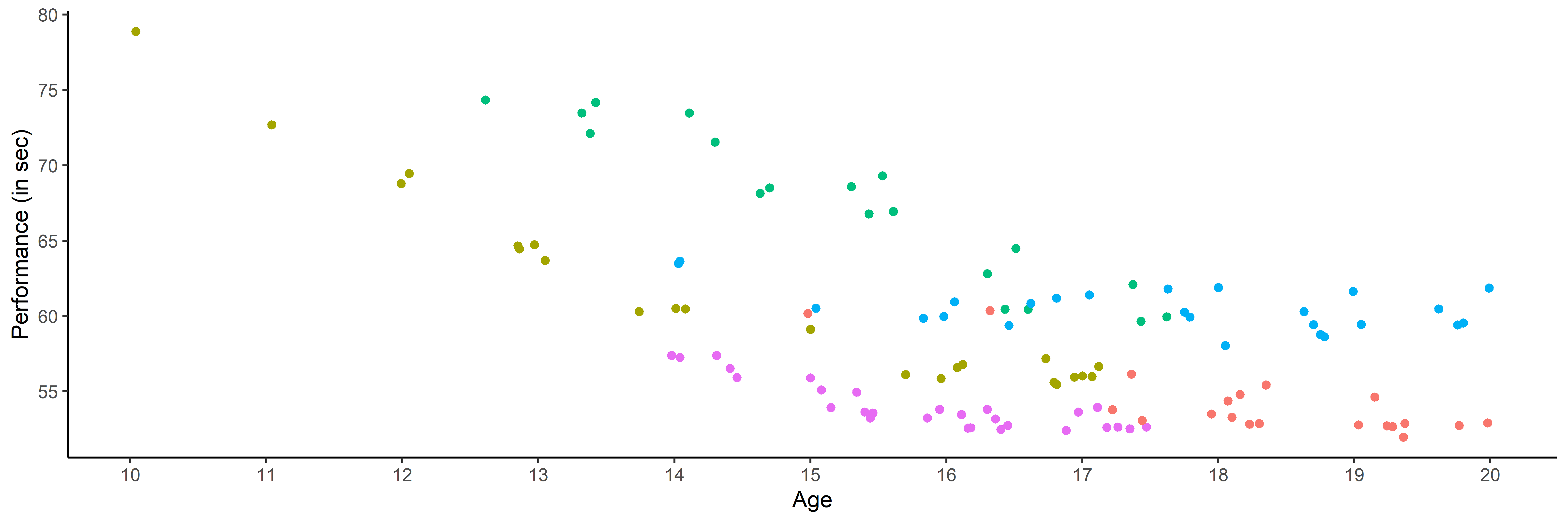}
       \caption{Time series representing the evolution of performances in 100m freestyle competitions between 10 and 20 years for 5 different swimmers, differentiated by colors.} 
		\label{illu_data_swimming}
    \end{center}
\end{figure*}

Before diving into modelling considerations, let us provide a motivational example used throughout the article to illustrate the kind of problems we expect to tackle. Assume that we have observed results of swimming competitions for thousands of athletes from 10 to 20 years old. In order to improve talent detection, one might be interested in using those data to forecast future performances for new young swimmers (e.g observed only between 10 and 14 years old). Such a data set is composed of thousands of irregular age-performance time series, where each swimmer would have specific number and locations of their data points, as illustrated in \Cref{illu_data_swimming}. These examples come from a real data set, thoroughly described in \Cref{sec:exp}, which was originally the applicative motivation for developing methods described in the present paper. The following multi-task GPs framework is tailored to simultaneously allocate individuals into clusters while learning model parameters, and then provide probabilistic predictions for future performances of any young swimmer, by sharing information between resembling individuals through common mean trends.

\subsection{Notation}
To remain consistent with this illustrative example, we refer throughout to the input variables as \emph{timestamps} and use the term \emph{individual} as a synonym of batch or task. 
However, although the temporal formulation helps intuition, the present framework still applies to the wide range of inputs one can usually think of in GP applications.
As we suppose the data set to be composed of point-wise observations from multiple functions, the set of all indices is denoted by $\I \subset \mathbb{N}$, which in particular contains $\{ 1, \dots, M\}$, the indices of the observed individuals (i.e. the training set).
Since the input values are defined over a continuum, let us name $\T$ this input space (we can assume  $\T \subset \mathbb{R}$ here for simplicity). 
Moreover, since the following model is defined for clustering purposes, the set of indices $\K = \{ 1,\dots, K \}$ refers to the $K$ different groups of individuals.
For the sake of concision, the notation is shortened as follows: for any object $x$, $\{x_i\}_i = \{x_1, \dots, x_M \}$ and $\{x_k\}_k = \{x_1, \dots, x_K \}$.
\newline

We assume data is collected from $M$ different sources, such that a set of $N_i$ input-output values $\acc{ \paren{t_i^1, y_i(t_i^1) }, \dots, \paren{t_i^{N_i}, y_i(t_i^{N_i}) } }$ constitutes the observations for the $i$-th individual. 
Below follows additional convenient notation:

\begin{itemize}
  \item $\ti = \{ t_i^1,\dots,t_i^{N_i} \}$, the set of timestamps for the $i$-th individual,
  \item $\yi = y_i(\ti)$, the vector of outputs for the $i$-th individual,
  \item $\Ut = \bigcup\limits_{i = 1}^M \ti$, the pooled set of all timestamps among individuals,
  \item $N = card(\Ut)$, the total number of observed timestamps.
\end{itemize}

Let us stress that the input values may vary both in number and location among individuals, and we refer as a \emph{common grid} of timestamps to the case where $\ti =  \Ut, \ \forall i \in \I$.
Otherwise, we call it an \emph{uncommon grid}.
Besides, in order to define a GP mixture model, a latent binary random vector $Z_i = \paren{Z_{i1}, \dots, Z_{iK} }^{\intercal}$ needs to be associated with each individual, indicating in which cluster it belongs. 
Namely, if the $i$-th individual comes from the $k$-th cluster, then $Z_{ik} = 1$ and 0 otherwise.
Moreover, we assume these latent variables to come from the same multinomial distribution: $Z_i \sim \mathcal{M}(1, \boldsymbol{\pi}), \ \forall i \in \I$, with a vector of mixing proportions $\boldsymbol{\pi} = \paren{\pi_1, \dots, \pi_K}^{\intercal}$ and $\sumk \pi_k = 1$.

\subsection{Model and Assumptions}
\label{sec:model}
Assuming that the $i$-th individual belongs to the $k$-th cluster, we can define its functional expression as the sum of a cluster-specific mean process and an individual-specific centred process:  
\begin{equation*}
y_i(t) = \mu_k(t) + f_i(t) + \varepsilon_i(t), \ \forall t \in \mathcal{T},
\end{equation*}
where: 
\begin{itemize}
  \item $\muk(\cdot) \sim \mathcal{GP} (\mk(\cdot), c_{\gammak}(\cdot,\cdot))$ is the common mean process of the $k$-th cluster,
  \item $f_i(\cdot) \sim \mathcal{GP} (0,\xi_{\theta_i}(\cdot,\cdot))$ is the specific process of the $i$-th individual,
  \item $\varepsilon_i(\cdot) \sim \mathcal{GP} (0,\sigma_i^2 I)$ is the error term.
\end{itemize}

This general model depends upon several mean and covariance parameters, fixed as modelling choices, and hyper-parameters to be estimated:

\begin{itemize}
	\item $\forall k \in \K, \ \mk(\cdot)$ is the prior mean function of the $k$-th cluster,
  \item $\forall k \in \K, \ c_{\gammak}(\cdot, \cdot)$ is the covariance kernel with hyper-parameters $\gammak$,
  \item $\forall i \in \I, \ \xi_{\thetai}(\cdot, \cdot)$ is the covariance kernel with hyper-parameters $\theta_i$,
  \item $\forall i \in \I, \ \sigma_i^2 \in \mathbb{R}$ is the noise variance associated with the $i$-th individual,
  \item $\forall i \in \I, \ $ we define the shorthand  $\psii(\cdot,\cdot) = \xi_{\theta_i}(\cdot,\cdot) + \sigma_i^2 I$, 
  \item $\Theta = \acc{\gammakk, \thetaii, \sigmaii, \pii }$, the set of all hyper-parameters of the model.
\end{itemize}

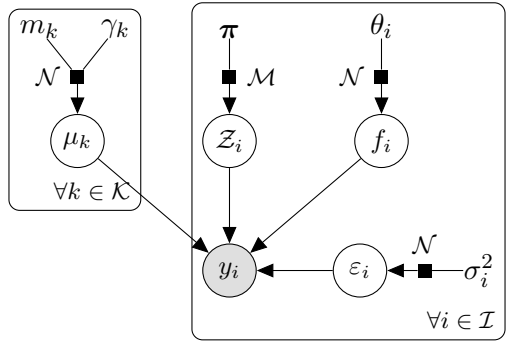
\begin{figure}
\begin{center}
\begin{tikzpicture}
  
  \node[obs]                               (y) {$y_i$};

  \node[latent, above=of y, xshift=-2cm](mu){$\mu_k$};
  \node[const, above=of mu, xshift=-0.5cm] (m) {$m_k$};
  \node[const, above=of mu, xshift= 0.5cm] (tk) {$\gamma_k$};

  \node[latent, above=of y, xshift= 0cm](z){$\mathcal{Z}_i$};
  \node[const, above=of z, xshift= 0cm]	   (pi) {$\pii$};

  \node[latent, above=of y, xshift=2cm] (f) {$f_i$};
  \node[const, above=of f, xshift= 0cm]	   (ti) {$\theta_i$};
  
  \node[latent, right= 1cm of y]           (e) {$\varepsilon_i$};
  \node[const, right= 1cm of e]		       (s) {$\sigma_i^2$};

  \factor[above=of mu] {mu-f} {left:$\mathcal{N}$} {m,tk} {mu} ;
  \factor[above=of f] {f-f} {left:$\mathcal{N}$} {ti} {f} ;
  \factor[above=of z] {z-f} {right:$\mathcal{M}$} {pi} {z} ;
  \factor[right=of e] {e-f} {above:$\mathcal{N}$} {s} {e} ;

  \edge {f,e,z,mu} {y} ;

  \plate {} {(f)(y)(e)(z)(ti)(s)} {$\forall i \in \mathcal{I}$} ;
  \plate {} {(mu)(m)(tk)} {$\forall k \in \K$} ;

  \end{tikzpicture}
  \caption{Graphical model of dependencies between variables in the multi-task Gaussian Processes mixture model.}
  \label{graph_model} 
  \end{center}     
  \end{figure}

 One can note that we assume here the error term to be individual-specific, although we could also assume it to be cluster-specific and thus indexed by $k$.
  Such a choice would result in a valid model since the upcoming developments remain tractable if we substitute $\varepsilon_k$ to $\varepsilon_i$ everywhere, and associate $\sigma_k^2 I$ with $c_{\gammak}(\cdot, \cdot)$ instead of $\xi_{\thetai}(\cdot, \cdot)$. 
  A discussion about additionally available assumptions on the covariance structures follows in \Cref{sec_hypo_hp}.
 In this paper, we seek an estimation for $\Theta$ among the above quantities, whereas the other objects are pre-specified in the model. 
  For instance, the prior mean $\mk(\cdot)$ is usually set to zero but could also integrate expert knowledge if available.
  Furthermore, we assume that:
  \begin{itemize}
   \item $\{\muk \}_k $ are independent,
   \item $\{ f_i \}_i$ are independent,
   \item $\{ \Zi \}_i$ are independent,
   \item $\{ \varepsilon_i \}_i$ are independent,
   \item $\forall i \in \I, \forall k \in \K, \ \mu_k$, $f_i$, $\Zi$, $\varepsilon_i$  and all pairwise combinations are independent.
 \end{itemize}
 We display a graphical model on \Cref{graph_model} to enlighten the relationships between the different components.
 From these hypotheses, we can naturally integrate out $f_i$ and derive the conditional prior distribution of $y_i(\cdot)$, providing a hierarchical formulation for the model:
 \begin{equation*}
 y_i(\cdot) \vert \{ \Zik = 1, \mu_k(\cdot) \} \sim \mathcal{GP} \paren{ \muk(\cdot), \psii(\cdot, \cdot)}, \ \forall i \in \I, \forall k \in \K.
 \end{equation*}

 As a consequence, the output processes $\{ y_i(\cdot) \vert \Zii, \{\mu_k(\cdot)\}_k \}_i$ are also independent (conditionally to the latent variables) from one another.
 Although this model is expressed in terms of infinite-dimensional GPs, we proceed to the inference using finite-dimensional sets of observations $\{\ti , \yi \}_i$.
 Therefore, we can write the joint conditional likelihood of the model (conditioning on the inputs is omitted throughout the paper for clarity):
 \begin{align*}
 p(\yii \vert \Zii, \{\mu_k(\Ut)\}_k, \thetaii, \sigmaii)
 &= \prodi  p(\yi \vert \Zi, \{\mu_k(\ti)\}_k, \theta_i, \sigma_i) \\
 &= \prodi \prodk  p(\yi \vert \Zik = 1, \mukti,  \theta_i, \sigma_i)^{\Zik} \\
 &= \prodi \prodk \N \paren{ \yi; \mukti, \Psiiti }^{\Zik},
 \end{align*}

 \noindent where  $\forall i \in \I, \ \Psiiti = \psii (\ti, \ti) = \left[ \psii(u, v) \right]_{u, v \in \ti}$ is a $N_i \times N_i$ covariance matrix.
 The mean processes being common to all individuals in a cluster, we need to evaluate their prior distributions on the pooled grid of timestamps $\Ut$: 
 \begin{align*}
 p(\mukkt \mid \gammakk ) 
 &= \prodk p(\mukt \mid \gammak )  \\ 
 &= \prodk \N \left(\mukt;  \mkt , \Ckt \right),
 \end{align*}
 \noindent where $\Ckt = c_{\gammak}(\Ut, \Ut) = \left[ c_{\gammak}(k, \ell) \right]_{k,l \in \Ut}$ is a $N \times N$ covariance matrix.  
 Finally, the joint distribution of the clustering latent variables also factorises over the individuals:
 \begin{align*}
 p(\Zii \mid \pii ) 
 &= \prodi p(\Zi \mid \pii )  \\
 &= \prodi \mathcal{M}(\Zi;1, \pii) \\
 &= \prodi \prodk \pik^{\Zik}.
 \end{align*}

 From all these expressions, the complete-data likelihood of the model can be derived: 
 \begin{align*}
 p(\yii, \Zii, \mukkt \mid  \Theta)
 &= p(\mukkt  \mid  \gammak) \prodi  p(\yi \mid \Zi, \mukkti , \thetai, \sigmai) p(\Zi \mid \pii)  \\
 &= \prodk \N \left(\mukt;  \mkt , \Ckt \right)  \prodi \paren{\pik \N \paren{ \yi; \mukti, \Psiiti }}^{\Zik} .
 \end{align*}

 This expression would usually serve to estimate the hyper-parameters $\Theta$, although it depends here on latent variables that cannot be evaluated directly. 
 Even if the prior distributions over $\Zii$ and $\mukkt$ are independent, the expressions of their respective posteriors would inevitably depend on each other.
 Nevertheless, it remains possible to derive variational approximations for these distributions that still factorise nicely over the terms $\Zi, \forall i \in \I$, and $\mukt, \forall k \in \K$.
 Consequently, the following inference procedure involves a variational EM algorithm that we shall detail after a quick discussion on the optional hypotheses for the model.

 \subsection{Assumptions on the Covariance Structure}
 \label{sec_hypo_hp}

 Throughout this paper, we detail a common ground procedure that remains consistent regardless of the covariance structure of the considered GPs. 
Notice that we chose a parametric distinction of the covariance kernels through the definition of hyper-parameters, different from one individual to another.
However, there are no theoretical restrictions on the underlying form of the considered kernels, and we indicate a differentiation on the sole hyper-parameters merely for convenience in writing.
A usual kernel in the GP literature is known as the \emph{exponentiated quadratic} kernel (also called sometimes squared exponential or radial basis function kernel).
This kernel only depends upon two hyper-parameters $\theta = \acc{v, \ell}$ such as: 

\begin{equation}
\label{eq:kernel}
k_{\mathrm{EQ}}\left(x, x^{\prime}\right)= v^{2} \exp \left(-\frac{\left(x-x^{\prime}\right)^{2}}{2 \ell^{2}}\right).
\end{equation}
Although beyond the scope of the present paper, let us mention the existence of a rich literature on kernel design, properties and combinations: see \citet[Chapter~4]{RasmussenGaussianprocessesmachine2006a} or \cite{DuvenaudAutomaticmodelconstruction2014} for comprehensive studies.
\newline

In the algorithm \algo \citep{LeroyMAGMAInferencePrediction2020}, the multi-task aspect is mainly supported by the mean process, although the model also allows information sharing among individuals through the covariance structure.
These two aspects being constructed independently, we could think of the model as potentially \emph{double multi-task}, both in mean and covariance.
More precisely, if we assume $\{ \thetaii, \sigmaii \} = \{ \theta_0, \sigma_0^2 \}, \ \forall i \in \I$, then all $f_i$ are assumed to be different realisations of the same GP, and thus all individuals contributes to the estimation of the common hyper-parameters.
Hence, such an assumption that may appear restrictive at first glance actually offers a valuable way to share common patterns between tasks.
Furthermore, the same kind of hypothesis can be proposed at the cluster level with $ \gammakk = \gamma_0, \ \forall k \in \K$.
In this case, we would assume that all the clusters' mean processes $\{\muk\}_k$ share the same covariance structure.
This property would indicate that the patterns, or the variations of the curves, are expected to be roughly identical from one cluster to another and that the differentiation would be mainly due to the mean values. 
Conversely, different covariance structures across kernels offer additional flexibility for the clusters to differ both in position and in trend, smoothness, or any property that could be coded in a kernel.
Speaking rather loosely, we may think of these different settings as a trade-off between flexibility and information sharing, or as a choice between an individual or collective modelling of the covariance.  
Overall, our algorithm provides 4 different settings, offering a rather wide range of assumptions for an adequate adaptation to different applicative situations.
Note that the computational considerations are also of paramount importance when it comes to optimising a likelihood over a potentially high number of parameters.
Hence, we display on \Cref{table_hyp_hp} a summary of the 4 different settings, providing a shortening notation along with the associated number of hyper-parameters (or sets of hyper-parameters in the case of $\theta_i$ and $\gamma_k$) that are required to be learnt in practice. 

\begin{table}
\caption{Summary of the 4 available assumptions on the hyper-parameters, with their respective shortening notation and the associated number of sets of hyper-parameters (HPs) to optimise.}
\label{table_hyp_hp}
\begin{center}
\begin{tabular}{c|cc|cc|}
\cline{2-5}
& \multicolumn{2}{c|}{$\theta_0 = \theta_i, \forall i \in \I$} & \multicolumn{2}{c|}{$\theta_i \neq \theta_j, \forall i \neq j$} \\
& Notation                      & Nb of HPs                     & Notation                        & Nb of HPs                       \\ \hline
\multicolumn{1}{|c|}{$\gamma_0 = \gamma_k, \forall k \in  \K$}   & $\Hoo$                        & 2                             & $\Hoi$                          & M + 1                          \\ \cline{1-1}
\multicolumn{1}{|c|}{$\gamma_k \neq \gamma_l, \forall k \neq l$} & $\Hko$                        & K + 1                          & $\Hki$                          & M + K                          \\ \hline
\end{tabular}
\end{center}
\end{table}

\section{Inference}
\label{sec:inference}

Although a fully Bayesian point-of-view could be taken on the learning procedure by defining prior distributions of the hyper-parameters and directly use an MCMC algorithm \citep{RasmussenGaussianprocessesmachine2006a,YangSmoothingMeanCovariance2016} for approximate inference on the posteriors, this approach remains computationally challenging in practice. 
Conversely, variational methods have proved to be highly efficient to conduct inference in difficult GP problems \citep{TitsiasVariationalLearningInducing2009,HensmanGaussianProcessesBig2013} and may apply in our context as well.
By introducing an adequate independence assumption, we are able to derive a variational formulation leading to analytical approximations for the true hyper-posterior distributions of the latent variables. 
Then, these hyper-posterior updates allow the computation of a lower bound of the true log-likelihood, thereby specifying the E step of the VEM algorithm \citep{AttiasVariationalBaysianFramework2000} that conducts the overall inference.
Alternatively, we can maximise this lower bound with respect to the hyper-parameters in the M step for optimisation purpose, to provide estimates.
By iterating on these two steps until convergence (pseudo-code  in \Cref{algo_VEM}), the procedure is proved to reach local optima of the lower bound \citep{BoydConvexOptimization2004}, usually in a few iterations.
Regarding the previous illustrative example, this learning step allocates swimmers presenting similar progression patterns into dedicated clusters while simultaneously estimating the underlying mean trend of those clusters (E step) and optimising all hyper-parameters controlling each specific covariance structure (M step).
For the sake of clarity, the shorthand $\Z = \Zii$ and $\mub = \mukkt$ is used in this section when referring to the corresponding set of latent variables.
\subsection{Variational EM Algorithm}
\label{sec:VEM_algo}

We seek an appropriate and analytical approximation $q(\Z, \mub)$ for the exact hyper-posterior distribution $p(\Z, \mub \mid \yii, \Theta)$. 
Let us first notice that for any distribution $q(\Z, \mub)$, the following decomposition holds for the observed-data log-likelihood:
\begin{equation}
\label{eq_VEM}
	\log p(\yii \mid \Theta) = \mathrm{KL} \paren{ q \| p } + \mathcal{L}(q; \Theta),
\end{equation}
\noindent with: 
\begin{align*}
	\mathrm{KL} \paren{ q \| p } &= \int \int q(\Z, \mub) \log \frac{q(\Z, \mub)}{p(\Z , \mub \mid \yii, \Theta)} \dif \Z \dif \mub, \\
	\mathcal{L}(q; \Theta) &= - \int \int q(\Z, \mub) \log \frac{q(\Z, \mub)}{p(\Z , \mub , \yii \mid  \Theta)} \dif \Z \dif \mub.
\end{align*}
Therefore, we expressed the intractable log-likelihood of the model by introducing the Kullback-Leibler (KL) divergence between the approximation $q(\Z, \mub)$ and the corresponding true distribution $p(\Z , \mub \mid \yii , \Theta)$.
The right-hand term $\mathcal{L}(q; \Theta)$ in \eqref{eq_VEM} defines a so-called \emph{lower bound} for $\log p(\yii \mid \Theta)$ since a KL divergence is nonnegative by definition.
This lower bound depends both upon the approximate distribution $q(\cdot)$ and the hyper-parameters $\Theta$, while remaining tractable under adequate assumptions.
By maximising $\mathcal{L}(q; \Theta)$ alternatively with respect to both quantities, optima for the hyper-parameters shall be reached.
To achieve such a procedure, the following factorisation is assumed for the approximated distribution:
\begin{equation*}
	q(\Z, \mu) = q_{\Z}(\Z) q_{\mub}(\mub).
\end{equation*} 

Colloquially, we could say that the independence property that lacks to compute explicit hyper-posterior distributions is \emph{imposed}. 
Such a condition restricts the family of distributions from which we choose $q(\cdot)$, and we now seek approximations within this family that are as close as possible to the true hyper-posteriors.

\paragraph{E step \\}

In the expectation step (E step) of the VEM algorithm, the lower bound of the marginal likelihood $\mathcal{L}(q; \Theta)$ is maximised with respect to the distribution $q(\cdot)$, considering that initial or previously estimated values for $\hat{\Theta}$ are available.
Making use of the factorised form previously assumed, we can derive analytical expressions for the optimal distributions over $q_{\Z}(\Z)$ and $q_{\mub}(\mub)$.
As the computing of each distribution involves taking an expectation with respect to the other one, this suggests an iterative procedure where whether the initialisation or a previous estimation serves in the current optimisation process.
Therefore, two propositions are introduced below, respectively detailing the exact derivation of the optimal distributions $\hat{q}_{\Z}(\Z)$ and $\hat{q}_{\mub}(\mub)$ (all proofs are deferred to the corresponding \Cref{sec:proofs}).

\begin{proposition}
\label[proposition]{prop:E_step_Z}
Assume that the hyper-parameters $\hat{\Theta}$ and the variational distribution $\hat{q}_{\mub}(\mub) = \prodk \N \paren{\mu_k(\Ut) ; \mkhat(\Ut) , \Chatkt}$ are known. 
The optimal variational approximation $\hat{q}_{\Z}(\Z)$ of the true hyper-posterior $p\paren{\Z \vert \yii, \hat{\Theta}}$ factorises as a product of multinomial distributions: 
	\begin{equation}
	\label{eq_var_Z}
		\hat{q}_{\Z}(\Z) = \prodi \mathcal{M} \paren{\Zi ; 1 , \taui =  \paren{ \uptau_{i1}, \dots, \uptau_{iN} }^{\intercal} },
	\end{equation}
	
	\noindent where: 
	\begin{equation}
		\tauik = \dfrac{\pikhat \ \N \paren{\yi ; \mkhat(\ti) , \Psiihat^{\ti}} \exp\paren{ -\frac{1}{2} \tr{ {\Psiihat^{\ti}}^{-1}  \Chatkti } } }{\sum\limits_{l = 1}^{K} \pilhat \ \N \paren{\yi ; \hat{m}_l(\ti) , \Psiihat^{\ti}} \exp\paren{ -\frac{1}{2} \tr{ {\Psiihat^{\ti}}^{-1}  \Chatlti } }}, \ \forall i \in \I, \forall k \in \K.
	\end{equation}
\end{proposition}

\begin{proposition}
\label[proposition]{prop:E_step_mu}
Assume that the hyper-parameters $\hat{\Theta}$ and the variational distribution $\hat{q}_{\Z}(\Z) = \prodi \mathcal{M} \paren{\Zi ; 1 , \taui}$ are known. 
The optimal variational approximation $\hat{q}_{\mub}(\mub)$ of the true hyper-posterior $p\paren{\mub \vert \yii, \hat{\Theta}}$ factorises as a product of multivariate Gaussian distributions: 
	\begin{equation}
	\label{eq_var_mu}
		\hat{q}_{\mub}(\mub) = \prodk \N \paren{\mu_k(\Ut) ; \mkhat(\Ut) , \Chatkt},
	\end{equation}
	with: 
	\begin{itemize}
		\item $\Chatkt = \paren{ \Ckhatt^{-1} + \sumi \tauik \Psitilde^{-1}}^{-1}, \ \forall k \in \K$, 
		\item $\mkhat(\Ut) = \Chatkt \paren{ \Ckhatt^{-1} \mkt + \sumi \tauik \Psitilde^{-1} \tilde{\mathbf{y}}_i },  \ \forall k \in \K$,
	\end{itemize}
	
	\noindent where the following shorthand notation is used: 
		\begin{itemize}
		\item $\tilde{\mathbf{y}}_i = \paren{\mathds{1}_{ [t \in \ti ]} \times y_i(t)}_{t \in \Ut}$ ($N$-dimensional vector),
		\item $\Psitilde = \croch{ \mathds{1}_{ [t, t' \in \ti]} \times \psiihat \paren{t, t'} }_{t, t' \in \Ut}$ ($N \times N$ matrix).
		\end{itemize}
\end{proposition}

Notice that the \emph{forced} factorisation we assumed between $\Z$ and $\mub$ for approximation purpose additionally offers an induced independence between individuals as indicated by the factorisation in \eqref{eq_var_Z}, and between clusters in \eqref{eq_var_mu}.

\paragraph{M step \\}
At this point, we have fixed an estimation for $q(\cdot)$ in the lower bound that shall serve to handle the maximisation of $\mathcal{L}(\hat{q}, \Theta)$ with respect to the hyper-parameters.
This maximisation step (M step) depends on the initial assumptions on the generative model (\Cref{table_hyp_hp}), resulting in four different versions for the VEM algorithm (the E step is common to all of them, the branching point is here).
\begin{proposition}
\label[proposition]{prop:M_step}
Assume the variational distributions $\hat{q}_{\Z}(\Z) = \prodi \mathcal{M} \paren{\Zi ; 1 , \taui}$ and $\hat{q}_{\mub}(\mub) = \prodk \N \paren{\mu_k(\Ut) ; \mkhat(\Ut) , \Chatkt}$ to be known.
For a set of hyper-parameters  $\Theta = \{ \gammakk, \thetaii,  \sigmaii, \pii \}$, the optimal values are given by: 
	\begin{equation*}
		\hat{\Theta} = \argmax\limits_{\Theta} \mathbb{E}_{ \{ \Z, \mub \} } \croch{ \log p(\yii, \Z , \mub \vert \Theta) }, 
	\end{equation*}
	
\noindent where $\mathbb{E}_{ \{ \Z, \mub \} }$ indicates an expectation taken with respect to $\hat{q}_{\mub}(\mub)$ and $\hat{q}_{\Z}(\Z)$.
	In particular, optimal values for $\pii$ can be computed explicitly with:
	
	\begin{equation*}
		\pikhat = \dfrac{1}{M} \sumi \tauik, \ \forall k \in \K.
	\end{equation*}
\noindent The remaining hyper-parameters are estimated by numerically solving the following maximisation problems, according to the situation.
	Let us note:
	\begin{align*}
		\mathcal{L}_k \paren{ \mathbf{x}; \mathbf{m}, S } &= \log \mathcal{N} \paren{\mathbf{x}; \mathbf{m}, S } - \dfrac{1}{2} \tr{\Chatkt {S}^{-1}}, \\
		\mathcal{L}_i \paren{ \mathbf{x}; \mathbf{m}, S } &= \sumk \tauik \paren{ \log \mathcal{N} \paren{\mathbf{x}; \mathbf{m}, S } - \dfrac{1}{2} \tr{\Chatkti {S}^{-1}} }. 
	\end{align*}	
	Then, for hypothesis $\Hki$: 		
	\begin{itemize}
		\item $\hat{\gamma}_k = \argmax\limits_{\gammak} \mathcal{L}_k \paren{\mkhat(\Ut); m_k(\Ut), \Ckt }, \ \forall k \in \K$,
		\item $( \hat{\theta}_i, \hat{\sigma}_i^2 ) = \argmax\limits_{\thetai, \sigmai} \mathcal{L}_i \paren{ \yi; \mkhat(\ti), \Psiiti }, \ \forall i \in \I $.
	\end{itemize}
	For hypothesis $\Hko$: 		
	\begin{itemize}
		\item $\hat{\gamma}_k = \argmax\limits_{\gammak} \mathcal{L}_k \paren{\mkhat(\Ut); m_k(\Ut), \Ckt }, \ \forall k \in \K$,
		\item $( \hat{\theta}_0, \hat{\sigma}_0^2 ) = \argmax\limits_{\theta_0, \sigma_0^2} \sumi \mathcal{L}_i \paren{ \yi; \mkhat(\ti), \boldsymbol{\Psi}_{\theta_0, \sigma_0^2}^{\ti} } $.
	\end{itemize}
	For hypothesis $\Hoi$: 		
	\begin{itemize}
		\item $\hat{\gamma}_0 = \argmax\limits_{\gamma_0} \sumk \mathcal{L}_k \paren{\mkhat(\Ut); m_k(\Ut), \mathbf{C}_{\gamma_0}^{\mathbf{t}} },$
		\item $( \hat{\theta}_i, \hat{\sigma}_i^2 ) = \argmax\limits_{\thetai, \sigmai}  \mathcal{L}_i \paren{ \yi; \mkhat(\ti), \Psiiti }, \ \forall i \in \I $.
	\end{itemize}
	For hypothesis $\Hoo$:
	\begin{itemize}
		\item $\hat{\gamma}_0 = \argmax\limits_{\gamma_0} \sumk \mathcal{L}_k \paren{\mkhat(\Ut); m_k(\Ut), \mathbf{C}_{\gamma_0}^{\mathbf{t}} }$,
		\item $( \hat{\theta}_0, \hat{\sigma}_0^2 ) = \argmax\limits_{\theta_0, \sigma_0^2} \sumi  \mathcal{L}_i \paren{ \yi; \mkhat(\ti), \boldsymbol{\Psi}_{\theta_0, \sigma_0^2}^{\ti} } $.
	\end{itemize}

\end{proposition}

Let us stress that, for each sub-case, explicit gradients are available for the functions to maximise, facilitating the optimisation process with gradient-based methods \citep{HestenesMethodsconjugategradients1952,BengioGradientBasedOptimizationHyperparameters2000}.
The current version of our code implements those gradients and makes use of them within the L-BFGS-B algorithm \citep{NocedalUpdatingquasiNewtonmatrices1980,MoralesRemarkalgorithmLBFGSB2011} devoted to the numerical maximisation.
As previously discussed, the hypothesis $\Hki$ necessitates to learn $M + K$ sets of hyper-parameters.
However, we notice in \Cref{prop:M_step} that the factorised forms defined as the sum of a Gaussian log-likelihoods and trace terms offer a way to operate the maximisations in parallel on simple functions.
Conversely, for the hypothesis $\Hoo$, only $2$ sets of hyper-parameters need to be optimised, namely $\gamma_0$, and $\{\theta_0,\sigma_0^2 \}$.
The small number of functions to maximise is explained by the fact that they are defined as larger sums over all individuals (respectively all clusters).
Moreover, this context highlights a multi-task pattern in covariance structures, since each individual (respectively cluster) contributes to the learning of shared hyper-parameters. 
In practice, $\Hoo$ is far easier to manage, and we generally reach robust optima in a few iterations.
On the contrary, the settings with many hyper-parameters to learn, using mechanically less data for each, may lead more often to computational burden or pathological results.  
The remaining hypotheses, $\Hoi$ and $\Hko$, are somehow middle ground situations between the two extremes and might be used as a compromise according to the problem being dealt with.

\subsection{Initialisation}
\label{sec:initialisation}

Below some modelling choices are discussed, in particular the initialisation of some quantities involved in the VEM algorithm:

\begin{itemize}
\item $\{m_k(\cdot)\}_k$; the mean functions from the hyper-prior distributions of the associated mean processes $\{\mu_k(\cdot)\}_k$.
As it may be difficult to pre-specify meaningful values in the absence of external or expert knowledge, these values are often assumed to be $0$. However, it remains possible to integrate information in the model by this mean. 
However, as exhibited in \Cref{prop:E_step_mu}, the influence of $\{m_k(\cdot)\}_k$ in hyper-posterior computations decreases rapidly when $M$ grows in a multi-task framework.
\item $\gammakk$, $\thetaii$ and $\sigmaii$; the kernel hyper-parameters. 
We already discussed that the form itself of kernels has to be chosen as well, but once set, we would advise initiating $\gammakk$ and ${\thetaii}$ with close and reasonable values whenever possible.
As usual in GP models, nearly singular covariance matrices and numerical instability may occur for pathological initialisations, in particular for the hypotheses, like $\Hki$, with many hyper-parameters to learn.
This behaviour frequently occurs in the GP framework, and one way to handle this issue is to add a so-called \emph{jitter} term \citep{BernardoRegressionclassificationusing1998} on the diagonal of the ill-defined covariance matrices.
\item $\{\tauik\}_{ik}$; the estimated individual membership probabilities (or $\pii$; the prior vector of clusters' proportions).
Both quantities are valid initialisation depending on whether we start the VEM iterations by an E step or an M step.
If we only want to set the initial proportions of each cluster in the absence of additional information, we may merely specify $\pii$ and start with an E step.
Otherwise, if we insert the results from a previous clustering algorithm as an initialisation, the probabilities $\tauik$ for each individual and cluster can be fully specified before proceeding to an M step (or to the $\hat{q}_{\mub}(\mub)$'s computing and then the M step). 
\end{itemize}

Let us finally stress that the convergence (to local maxima) of VEM algorithms partly depends on these initialisations.
Different strategies have been proposed in the literature to manage this issue, among which simulated annealing \citep{UedaDeterministicannealingEM1998} or repeated short runs \citep{BiernackiChoosingstartingvalues2003}.

\subsection{Pseudocode}
\label{sec:pseudo_code}

The overall algorithm is called \algoclust (as an extension of the algorithm \algo to cluster-specific mean GPs) and we provide below the pseudo-code summarising the inference procedure.
The corresponding R code is available at \url{https://github.com/ArthurLeroy/MAGMAclust}.

\begin{algorithm}
\caption{\algoclust: Variational EM algorithm}
\label{algo_VEM}
\begin{algorithmic}
\STATE Initialise $\{m_k(\Ut) \}_k$, $\Theta = \acc{ \gammakk , \thetaii , \sigmaii}$ and $\{ \taui^{ini} \}_i$ (or $\pii$).
\WHILE{not converged} 
\STATE E step: Optimise $\mathcal{L}(q; \Theta)$ w.r.t. $q(\cdot)$:
\STATE \hspace{1.1cm} $\hat{q}_{\Z }(\Z) = \prodi \mathcal{M}(\Zi; 1, \taui).$ 
\STATE \hspace{1.1cm} $\hat{q}{\mub}(\mub) = \prodk \mathcal{N}(\mukt; \mkhat(\Ut), \Chatkt).$ 
\newline

\STATE M step: Optimise $\mathcal{L}(q; \Theta)$ w.r.t. $\Theta$:
\STATE \hspace{1.1cm} $\hat{\Theta} = \argmax\limits_{\Theta} \mathbb{E}_{\Z, \mub } \croch{ \log p(\yii, \Z, \mub \vert \Theta) } .$
\ENDWHILE
\RETURN $\hat{\Theta}$, $\{\taui\}_i$ $\{\mkhat(\Ut)\}_k$, $\{\Chatkt\}_k$.
\end{algorithmic}
\end{algorithm}

\subsection{Model Selection}
\label{sec:model_selection}
The question of the adequate choice of $K$ in clustering applications is a recurrent concern in practice.
Many criteria have been introduced in the literature, among which those relying on penalisation of the likelihood like AIC \citep{Akaikenewlookstatistical1974} or BIC \citep{SchwarzEstimatingDimensionModel1978} for instance.
Whereas we seek a BIC-like formula, we can recall that the likelihood $p(\yii \mid \hat{\Theta})$ cannot be computed directly in the present context.
However, as for inference, we may still use the previously introduced lower bound $\mathcal{L}(\hat{q}; \hat{\Theta})$ to adapt a so-called variational-BIC (VBIC) quantity to maximise, as proposed in \cite{YouVariationalBayesEstimation2014}.
The expression of this criterion is provided below while we defer the full derivation of the lower bound to \Cref{sec:proof_BIC}.

\begin{proposition}
\label[proposition]{prop:BIC}

After convergence of the VEM algorithm, a variational-BIC expression can be derived as:
\begin{align*}
	BIC_{var}(K) 
	&= \mathcal{L}(\hat{q}; \hat{\Theta}) - \dfrac{\mathrm{card}\{HP \}}{2} \log M \\
	&= \sumi \sumk  \croch{  \tauik \paren{ \log \mathcal{N}\paren{ \yi; \mkhat(\ti), \Psiihatti } - \dfrac{1}{2}  \tr{ \Chatkt \Psiihatti^{-1}} + \log \dfrac{\hat{\pik}}{\tauik}  } } \\
	& \hspace{0.5cm} + \sumk \Bigg[ \log \mathcal{N} \paren{ \mkhat(\Ut); \mkt , \Ckhatt }  - \dfrac{1}{2} \tr{ \Chatkt \Ckhatt^{-1}}  \\ 
	& \hspace{2cm} + \dfrac{1}{2} \log \abs{ \Chatkt } + N \log 2 \pi +  N  \Bigg] - \dfrac{\alpha_i + \alpha_k + (K - 1)}{2} \log M,
\end{align*}

\noindent where:
\begin{itemize}
	\item $\alpha_i$ is the number of hyper-parameters from the individual processes' kernels,
	\item $\alpha_k$ is the number of hyper-parameters from the mean processes' kernels,
	\item $K - 1$ is the number of free parameters $\hat{\pi_k}$ (because of the constraint $\sumk \pi_k = 1$ ).
\end{itemize}
\end{proposition}

Let us mention that the numbers $\alpha_i$ and $\alpha_k$ in the penalty term vary according to the considered modelling hypothesis ($\Hoo$, $\Hko$, $\Hoi$ or $\Hki$), see \Cref{table_hyp_hp} for details.

\section{Prediction}
\label{sec:prediction}

At this point, we would consider that the inference on the model is completed, since the training data set of observed individuals $\yii$ enabled to estimate the desired hyper-parameters and the distributions of latent variables.
For the sake of concision, we thus omit the writing of conditionings over $\hat{\Theta}$ in the sequel. 
Recalling our illustrative example, we would have used competition's results over a long period from thousands of swimmers for training the model, and we now expect to make predictions of future performances for any young athlete in the early stages of their career.
Therefore, let us assume the partial observation of a new individual, denoted by the index $*$, for whom we collected a few data points $y_*(\tst)$ at timestamps $\tst$.
Defining a multi-task GPs mixture prediction consists in seeking an analytical distribution $p(y_*(\cdot) \mid y_*(\tst), \yii)$, according to the information brought by: its own observations; the training data set; the cluster structure among individuals.  
As we aim at studying the output values $y_*(\cdot)$ at arbitrarily chosen timestamps, say $\tpred$ (the index $p$ stands for \emph{prediction}), we propose a new notation for the pooled vector of timestamps $\tpst = \begin{bmatrix}
\tpred \\
\tst
\end{bmatrix}$. 
This vector serves as a working grid on which the different distributions involved in the prediction procedure are evaluated.
In the absence of external restrictions, we would strongly advise to include the observed timestamps of all training individuals, $\Ut$, within $\tpst$, since evaluating the processes at these locations allows for sharing information across tasks.
Otherwise, any data points defined on timestamps outside of the working grid would be discarded from the multi-task aspect of the model.
In particular, if $\tpst = \Ut$, we may even use directly the variational distribution $q_{\mub}(\mub)$ computed in the VEM algorithm, and thus skip one step of the prediction procedure that is described below.
Throughout the section, we aim at defining a probabilistic prediction for this new individual, accounting for the information of all training data $\yii$.
To this end, we manipulate several distributions of the type $p(\cdot \mid \yii)$ and refer to them with the adjective \emph{multi-task}.
Additionally to highlighting the information-sharing aspect across individuals, this term allows us to distinguish the role of $\yii$ from the one of the newly observed data $y_*(\tst)$, which are now the reference data for establishing if a distribution is called a \emph{prior} or a \emph{posterior}. 
Deriving a predictive distribution in our multi-task GP framework requires to complete the following steps.

\begin{enumerate}
	\item Compute the hyper-posterior approximation of $\mukk$ at $\tpst$: $\hat{q}_{\mu}(\mukktpst)$,
	\item Deduce the multi-task prior distribution: $p(\yst (\tpst) \vert \Zst, \yii)$,
	\item Compute the new hyper-parameters $\{\thetast, \sigmast\}$ and $p(\Zst \mid \yst(\tst), \yii)$ via an EM,
	\item[3bis.]  Assign $\thetast = \theta_0$, $\sigmast = \sigma_0^2$ and compute directly $p(\Zst \vert \yst (\tst), \yii )$,
	\item Compute the multi-task posterior distribution: $p(\yst (\tpred) \vert \yst(\tst), \Zst, \yii )$,
	\item Deduce the multi-task GPs mixture prediction: $p(\yst (\tpred) \vert \yst(\tst), \yii )$.
\end{enumerate}

We already discussed the influence of the initial modelling hypotheses on the overall procedure.
Hence, let us display in \Cref{tab_pred_recap} a quick reminder helping to keep track of which steps need to be performed in each context.

\begin{table}
\begin{center}
\caption{Summary of the different steps to perform in the prediction procedure, according to the model assumptions and the target grid of timestamps.}
\label{tab_pred_recap}
\begin{tabular}{c|c|c|}
\cline{2-3}
& $\tpst = \Ut$ & $\tpst \neq \Ut$ \\ \hline
\multicolumn{1}{|c|}{$\Hoo$} & 2-3bis-4-5    & 1-2-3bis-4-5     \\ \cline{1-1}
\multicolumn{1}{|c|}{$\Hko$} & 2-3bis-4-5    & 1-2-3bis-4-5     \\ \cline{1-1}
\multicolumn{1}{|c|}{$\Hoi$} & 2-3-4-5       & 1-2-3-4-5        \\ \cline{1-1}
\multicolumn{1}{|c|}{$\Hki$} & 2-3-4-5       & 1-2-3-4-5        \\ \hline
\end{tabular}
\end{center}
\end{table}

\subsection{Posterior Inference on the Mean Processes}

In order to integrate the information contained in the shared mean processes, we first need to re-compute the variational approximation of $\mukk$'s hyper-posterior on the new $\tilde{N}$-dimensional working grid $\tpst$. 
By using once more \Cref{prop:E_step_mu}, it appears straightforward to derive this quantity that still factorises as a product of Gaussian distributions where we merely substitute the values of timestamps:

\begin{equation*}
\hat{q}_{\mub}(\mukktpst) = \prodk \N \paren{\mu_k(\tpst) ; \mkhat(\tpst) , \Chatktpst},
\end{equation*}
with: 
\begin{itemize}
	\item $\Chatktpst = \paren{ {\Ckhattpst}^{-1} + \sumi \tauik \Psitilde^{-1} }^{-1}, \ \forall k \in \K$, 
	\item $\mkhat(\tpst) = \Chatktpst \paren{ {\Ckhattpst}^{-1} m_k(\tpst) + \sumi \tauik {\Psitilde }^{-1} \tilde{\mathbf{y}}_i},  \ \forall k \in \K$,
\end{itemize}
\noindent where the following shorthand notation is used: 
\begin{itemize}
	\item $\tilde{\mathbf{y}}_i = \paren{\mathds{1}_{ [t \in \ti ]} \times y_i(t)}_{t \in \tpst}$ ($\tilde{N}$-dimensional vector),
	\item $\Psitilde = \croch{ \mathds{1}_{ [t, t' \in \ti]} \times \psiihat \paren{t, t'} }_{t, t' \in \tpst}$ ($\tilde{N} \times \tilde{N}$ matrix).
\end{itemize}

We acknowledge that the subsequent analytical developments party rely on this variational approximate distribution $\hat{q}_{\mub}(\mukktpst)$, and may thus be considered, in a sense, as approximated as well.
However, this quantity provides a valuable closed-form expression that we can substitute to the true hyper-posterior in \Cref{prop_integrate_mu} below.

\subsection{Computation of the Multi-Task Prior Distributions}
For a sake of completeness, let us recall the equivalence between two ways of writing conditional distributions that are used in the subsequent results:

\begin{equation*}
p(\cdot \mid \Zst) = \prodk p(\cdot \mid \Zstk = 1)^{\Zstk}.
\end{equation*}

\noindent We may regularly substitute one to the other in the sequel depending on the handier in the context. 
Once the mean processes' distributions are re-computed on the working grid, their underlying influence shall be directly plugged into a marginalised multi-task prior over $y_*(\tpst)$ by integrating out the $\mukktpst$.
As the mean processes vanish, the new individual's outputs $y_*(\tpst)$ directly depends upon the training data set $\yii$, as highlighted in the proposition below.

\begin{proposition}
\label[proposition]{prop_integrate_mu}
For a set of timestamps $\tpst$, the multi-task prior distribution of $y_*$ knowing its clustering latent variable is given by: 
\begin{equation}
\label{eq_prior}
p(y_*(\tpst) \vert \Zst, \yii) = \prodk \mathcal{N} \paren{ y_*(\tpst); \mkhat(\tpst), \Chatktpst + \Psi_{\theta_*, \sigma_*^2}^{\tpst}}^{\Zstk}.
\end{equation}
\end{proposition}

\begin{proof}
Let us recall that, conditionally to their mean process, the individuals are independent of one another.
Then, for all $k \in \K$, we have:
\begin{align*}
p(y_*(\tpst) \vert \Zstk = 1, \yii) 
&= \int p(y_*(\tpst), \muk(\tpst) \vert \Zstk = 1, \yii) \dif \muk(\tpst) \\
&= \int p(y_*(\tpst) \vert \muk(\tpst),  \Zstk = 1) \underbrace{p(\muk(\tpst) \mid \Zstk = 1, \yii)}_{\approx q_{\mub}(\muk(\tpst))} \dif \muk(\tpst) \\
& \approx \int \N \paren{y_*(\tpst); \muk(\tpst), \Psi_{\theta_*, \sigma_*^2}^{\tpst} } \N \paren{\mu_k(\tpst) ; \mkhat(\tpst) , \Chatktpst} \dif \muk(\tpst) \\
&= \N \paren{y_*(\tpst); \mkhat(\tpst), \Chatktpst + \Psi_{\theta_*, \sigma_*^2}^{\tpst} }.
\end{align*}

The final line is obtained by remarking that such a convolution of Gaussian distributions remains Gaussian as well \citep[Chapter~2]{BishopPatternrecognitionmachine2006}, and we refer to \cite{LeroyMAGMAInferencePrediction2020} for the detailed calculus in this exact context. 
Therefore, we finally get: 

\begin{align*}
p(y_*(\tpst) \vert \Zst, \yii) 
&= \prodk p(y_*(\tpst) \vert \Zstk = 1, \yii)^{\Zstk} \\
&= \prodk \N \paren{y_*(\tpst); \mkhat(\tpst), \Chatktpst + \Psi_{\theta_*, \sigma_*^2}^{\tpst} }^{\Zstk}.
\end{align*}

\end{proof}

\subsection{Optimisation of the New Hyper-Parameters and Computation of the Clusters' Probabilities}
Now that the mean processes have been removed at the previous step, this section strongly resembles the classical learning procedure through an EM algorithm for a Gaussian mixture model.
In our case, it allows us both to estimate the hyper-parameters of the new individual $\{\theta_*, \sigma_*\}$ and to compute the hyper-posterior distribution of its latent clustering variable $\Zst$, which provides the associated clusters' membership probabilities $\taust$.
As before, E steps and M steps are alternatively processed until convergence, but this time by working with exact formulations instead of variational ones.

\paragraph{E step\\}

In the E step, hyper-parameters estimates are assumed to be known. 
Recalling that the latent clustering variable $\Zst$ is independent from the training data $\yii$, the multi-task hyper-posterior distribution maintains an explicit derivation:
\begin{align*}
p(\Zst \mid \yst(\tst), \yii, \thetasthat, \sigmasthat, \piihat) 
& \propto p(\yst(\tst) \mid \Zst, \yii, \thetasthat, \sigmasthat) p(\Zst \mid \piihat) \\
& \propto \prodk \acc{ \mathcal{N} \paren{ y_*(\tst); \mkhat(\tst), \Chatktst + \Psi_{\thetasthat, \sigmasthat}^{\tst}}^{\Zstk} } \prod\limits_{l =1}^K {\pilhat}^{Z_{*l}} \\
& \propto \prodk \paren{\pikhat \mathcal{N} \paren{ y_*(\tst); \mkhat(\tst), \Chatktst + \Psi_{\thetasthat, \sigmasthat}^{\tst}}}^{\Zstk}.  \\
\end{align*}

By inspection, we recognise the form of a multinomial distribution and thus retrieve the corresponding normalisation constant to deduce:

\begin{equation}
\label{eq_post_Z}
p(\Zst \mid \yst(\tst), \yii, \thetasthat, \sigmasthat, \piihat)  = \mathcal{M} \paren{\Zst; 1, \taust = (\uptau_{*1}, \dots, \uptau_{*K})^{\intercal}},
\end{equation}
\noindent with:
\begin{equation}
\label{taustk}
\taustk = \dfrac{\pikhat \N \paren{\yst(\tst) ; \mkhat(\tst) , \Chatktst + \Psi_{\thetasthat, \sigmasthat}^{\tst}}}{\sum\limits_{l = 1}^{K} \pilhat \N \paren{\yst(\tst) ; \hat{m}_l(\tst), \Chatltst + \Psi_{\thetasthat, \sigmasthat}^{\tst}}}, \ \forall k \in \K.
\end{equation}

\paragraph{M step\\}

Assuming to know the value of $\taust$, we may derive optimal values for the hyper-parameters of the new individual through the following maximisation:

\begin{equation*}
\{ \thetasthat, \sigmasthat \} = \argmax\limits_{\theta_*, \sigma_*} \mathbb{E}_{\Zst} \croch{ \log p(y_*(\tst), \Zst \mid \yii, \theta_*, \sigma_*, \piihat)}.
\end{equation*}

Let us note $ \mathcal{L}_*(\theta_*, \sigma_*)= \log p(y_*(\tst), \Zst \mid \yii, \theta_*, \sigma_*, \piihat)$.
By remarking that $\piihat$ has already been estimated previously, we may easily derive the expression to maximise with respect to $\theta_*$ and $\sigma_*$ in practice: 

\begin{align*}
\mathbb{E}_{\Zst} \croch{ \mathcal{L}_*(\theta_*, \sigma_*) }
&=\mathbb{E}_{\Zst} \croch{ \log p(\yst(\tst), \Zst \mid \yii,\theta_*, \sigma_*, \piihat} ) \\
&= \mathbb{E}_{\Zst} \croch{ \log p(\yst(\tst) \mid \Zst, \yii, \theta_*, \sigma_*) + \log p(\Zst \mid \piihat) } \\ 
&= \mathbb{E}_{\Zst} \croch{\log  \prodk \mathcal{N} \paren{ y_*(\tst); \mkhat(\tst), \Chatktst + \Psi_{\thetast, \sigmast}^{\tst}}^{\Zstk}} + C_1 \\
&= \sumk \mathbb{E}_{\Zst} \croch{\Zstk} \log \mathcal{N} \paren{ y_*(\tst); \mkhat(\tst), \Chatktst + \Psi_{\thetast, \sigmast}^{\tst}} + C_1 \\
&= \sumk \taustk \log \mathcal{N} \paren{ y_*(\tst); \mkhat(\tst), \Chatktst + \Psi_{\thetast, \sigmast}^{\tst}} + C_1,		
\end{align*}
where $C_1$ is a constant term. 
Thus, the optimisation in this case merely relies on the maximisation of a weighted sum of Gaussian log-likelihoods, for which gradients are well-known.  

\paragraph{3bis. \\}

In the case where the hyper-parameters are supposed to be common across individuals ($\Hoo$ or $\Hko$), there is no need to additional optimisation since we already have $\thetasthat = \hat{\theta}_0$ and $\sigmasthat = \hat{\sigma}_0$ by definition.
However, the probabilities of lying in each cluster $\taust$ for the new individual still need to be computed, which shall be handled by directly using the expression \eqref{taustk} from the E step.

\paragraph{3ter. \\}

Conversely, even if hyper-parameters for each individual are supposed to be different ($\Hoi$ or $\Hki$), it remains possible to avoid the implementation of an EM algorithm by stating $\taust = \piihat$. 
Such an assumption intuitively expresses that we would guess the membership probabilities of each cluster from the previously estimated mixing proportions, without taking new individual's observations into account. 
Although we would not recommend this choice for getting optimal results, it still seems to be worth a mention for applications with a compelling need to avoid EM's extra computations during the prediction process. 

\subsection{Computation of the Multi-Task Posterior Distributions}

Once the needed hyper-parameters have been estimated and the prior distribution established, the classical formula for GP predictions can be applied to the new individual, for each possible latent cluster.
First, let us recall the prior distribution by separating observed from target timestamps, and introducing a shorthand notation for the covariance:

\begin{equation*}
p(y_*(\tpst) \vert \Zstk = 1, \yii) = \mathcal{N} \paren{
\begin{bmatrix}
y_*(\tpred) \\
y_*(\tst)
\end{bmatrix};
\begin{bmatrix}
\mkhat(\tpred) \\
\mkhat(\tst)
\end{bmatrix}, 
\begin{pmatrix}
\boldsymbol{\Gamma}_k^{\tpred \tpred} & \boldsymbol{\Gamma}_k^{\tpred \tst} \\
\boldsymbol{\Gamma}_k^{\tst \tpred} & \boldsymbol{\Gamma}_k^{\tst \tst}
\end{pmatrix} }, \ \forall k \in \K, 
\end{equation*}

\noindent where $\boldsymbol{\Gamma}_k^{\tpred, \tpred} = \Chatk^{\tpred} + \Psi_{\theta_*, \sigma_*^2}^{\tpred}$ and likewise for the other blocks of the matrices.
Therefore, recalling that conditioning on the sub-vector of observed values $y_*(\tst)$ maintains a Gaussian distribution \citep{BishopPatternrecognitionmachine2006,RasmussenGaussianprocessesmachine2006a}, we can derive the multi-task posterior distribution for each latent cluster: 

\begin{equation}
\label{pred_clustspecific}
p( y_*(\tpred) \vert \Zstk = 1 ,y_*(\tst) , \yii) = \mathcal{N} \paren{ y_*(\tpred);\hat{\mu}_{*k}(\tpred), \boldsymbol{\widehat{\Gamma}}_{*k}^{\tpred} }, \ \forall k \in \K,
\end{equation}

\noindent where: 

\begin{itemize}
	\item $\hat{\mu}_{*k}(\tpred) = \mkhat(\tpred) + \boldsymbol{\Gamma}_k^{\tpred \tst} {\boldsymbol{\Gamma}_k^{\tst \tst}}^{-1} \paren{ y_*(\tst) - \mkhat(\tst)}, \ \forall k \in \K,$
	\item $\boldsymbol{\widehat{\Gamma}}_{*k}^{\tpred} = \boldsymbol{\Gamma}_k^{\tpred \tpred} - \boldsymbol{\Gamma}_k^{\tpred \tst} {\boldsymbol{\Gamma}_{k}^{\tst \tst}}^{-1} \boldsymbol{\Gamma}_k^{\tst \tpred}, \ \forall k \in \K.$
\end{itemize}

\subsection{Computation of the Multi-Task GPs Mixture Prediction}

To conclude, by summing over all possible combinations for the latent clustering variable $\Zst$, we can derive the final predictive distribution.

\begin{proposition}
\label[proposition]{prop_finalpred}
The multi-task GPs mixture posterior distribution for $y_*(\tpred)$ has the following form:

\begin{equation*}
p( y_*(\tpred) \mid y_*(\tst) , \yii) = \sumk \taustk \ \mathcal{N} \paren{ y_*(\tpred);\hat{\mu}_{*k}(\tpred), \boldsymbol{\widehat{\Gamma}}_{*k}^{\tpred}}.
\end{equation*}
\end{proposition}

\begin{proof}

Taking advantage of \eqref{pred_clustspecific} and the multi-task hyper-posterior distribution of $\Zst$ as computed in \eqref{eq_post_Z}, it is straightforward to integrate out the latent clustering variable: 
\begin{align*}
p( y_*(\tpred) \vert y_*(\tst) , \yii)
&= \sum\limits_{\Zst} p( y_*(\tpred), \Zst \vert y_*(\tst) , \yii) \\
&= \sum\limits_{\Zst} p( y_*(\tpred) \mid \Zst, y_*(\tst) , \yii) p(\Zst \mid y_*(\tst) , \yii)	\\
&= \sum\limits_{\Zst} \prodk \Big(\taustk \ p( y_*(\tpred) \mid \Zstk = 1 ,y_*(\tst) , \yii) \Big)^{\Zstk} \\
&= \sum\limits_{\Zst} \prodk \paren{ \taustk \ \mathcal{N} \paren{ y_*(\tpred);\hat{\mu}_{*k}(\tpred), \boldsymbol{\widehat{\Gamma}}_{*k}^{\tpred} }}^{\Zstk} \\
&= \sumk \taustk \ \mathcal{N} \paren{ y_*(\tpred);\hat{\mu}_{*k}(\tpred), \boldsymbol{\widehat{\Gamma}}_{*k}^{\tpred}},
\end{align*}

\noindent where we recall for the transition to the last line that $\Zstk = 1$ if the $*$-th individual belongs to the $k$-th cluster and $\Zstk = 0$ otherwise.
Hence, summing a product with only one non-zero exponent over all possible combination for $\Zst$ is equivalent to merely sum over the values of $k$, and the variable $\Zstk$ simply vanishes. 
\end{proof}

\paragraph{Alternative predictions \\}

Even though \Cref{prop_finalpred} provides an elegant probabilistic prediction in terms of GPs mixture, it remains important to notice that this quantity is no longer a Gaussian distribution.
In particular, the distribution of an output value at any point-wise evaluation is expected to differ significantly from a classical Gaussian variable, by being multi-modal for instance.
This property is especially true for individuals with high uncertainty about the clusters they probably belong to, whereas the distribution would be close to the a when $\taustk \approx 1$ for one cluster and almost zero for the others. 
While we believe that such a GPs mixture distribution highlights the uncertainty resulting from a possible cluster structure in data and offers a rather original view on the matter of GP predictions, some applications may suffer from this non-Gaussian final distribution.
Fortunately, it remains pretty straightforward to proceed to a simplification of the clustering inference by assuming that the $*$-individual only belongs to its more probable cluster, which is equivalent to postulate $max\{\taustk\}_k = 1$ and the others to be zero. 
In this case, the final Gaussian mixture turns back into a Gaussian distribution, and we retrieve a uni-modal prediction, easily displayed by its mean along with credible intervals. 

\section{Complexity Analysis for Training and Prediction}
\label{sec:complexity}

It is customary to stress that computational complexity is of paramount importance in GP models as a consequence of their usual cubic (resp. quadratic) cost in the number of data points for learning (resp. prediction). 
In the case of \algoclust, we use information from $M$ individuals scattered into $K$ clusters, each of them providing $N_i$ observations, and those quantities mainly specify the overall complexity of the algorithm. 
Moreover, $N$ refers to the number of distinct timestamps (i.e. $N \leq \sumi N_i$) in the training data set and corresponds to the dimension of the objects involved in the kernel-specific mean processes computations. 
Typically, the learning complexity would be proportional to one iteration of the VEM algorithm, which requires $\mathcal{O} \paren{M \times N_i^3 + K \times N^3 }$ operations. 

Let us stress that this complexity is linear in the number of tasks, which is significantly lower than classical multi-output GP approaches in the literature.
As detailed in \cite{AlvarezKernelsVectorValuedFunctions2012a}, algorithms that can be formulated as linear models of coregionalisations typically present a $\mathcal{O}(M^3 N^3)$ training complexity when not resorting to sparse approximations.
Several approximations have been developed to reduce this cost, for instance, by using pseudo-inputs to decrease the $N^3$ term, while lowering the $M^3$ term generally requires underlying independence assumptions \cite{AlvarezComputationallyEfficientConvolved2011a}. 
In the \emph{Experiments} section, the computational advantage of \algoclust is also empirically highlighted in \Cref{table_compare_pred}. 
\newline 

From a practical point of view, the hypotheses formulated on the hyper-parameters may influence the constant of this complexity but generally not in more than an order of magnitude.
For instance, the models under the assumption $\Hoo$ usually require less optimisation time in practice, although it does not change the number or the dimensions of the covariance matrices to inverse, which mainly control the overall computing time.
The dominating terms in this expression depend on the context, regarding the relative values of $M$, $N_i$, $N$ and $K$. 
In contexts where the number of individuals $M$ dominates, like with small common grids of timestamps for instance, the left-hand term would control the complexity, and clustering's additional cost would be negligible. 
Conversely, for a relatively low number of individuals or a large size $N$ for the pooled grid of timestamps, the right-hand term becomes the primary burden, and the computing time increases proportionally to the number of clusters compared to the original \algo algorithm.
\newline

During the prediction step, the re-computation of $\mukk$'s variational distributions implies $K$ inversions of covariance matrices with dimensions depending on the size of the prediction grid $\tpst$.
In practice though, if we fix a fine grid of target timestamps in advance, this operation can be assimilated to the learning step. 
In this case, the prediction complexity remains at most in the same order as the usual learning for a single-task GP, that is $\mathcal{O}(K \times N_*^3)$ (this corresponds to the estimation of the new individual's hyper-parameters, and would decrease to $\mathcal{O}(K \times N_*^2)$ for $\Hko$ or $\Hoo$).
We shall mention that the definition of a fine grid is generally desirable only for low-dimensional applications since we may quickly reach running time or memory limits as the input's dimension grows. 
In many contexts, most of the time-consuming learning steps can be performed in advance, and the immediate prediction cost for each new individual is negligible in comparison (generally comparable to a single-task GP prediction).

\section{Experiments}
\label{sec:exp}
The present section is dedicated to the evaluation of \algoclust on both synthetic and real data sets.
The performance of the algorithm is assessed in regards to its clustering and forecast abilities.
To this purpose, we introduce below the simulation scheme generating the synthetic data along with the measures used to compare our method to alternatives quantitatively.
Throughout, the \textit{exponentiated quadratic} (EQ) kernel, as defined in \Cref{eq:kernel}, serves as covariance structure for both generating data and modelling.
The manipulation of more sophisticated kernels remains a topic beyond the scope of the present paper, and the EQ proposes a fair common ground for comparison between methods.
Thereby, each kernel introduced in the sequel is associated with two hyper-parameters. 
Namely, $\ v \in \mathbb{R}^{+}$ represents a variance term whereas $\ell \in \mathbb{R}^{+}$ specifies the length-scale.
The synthetic data sets are generated following the general procedure below, with minor modifications according to the model assumptions $\Hoo$, $\Hko$, $\Hoi$ or $\Hki$: 

\begin{enumerate}
	\item Define a random working grid $\Ut \subset \croch{0,10}$ of $N = 200$ timestamps to study $M = 50$ individuals, scattered into $K$ clusters,
	\item Draw the prior mean functions for $\mukk$: $\mk(t) = at + b, \ \forall t \in \Ut, \forall k \in \K$, where $a \in \croch{-2, 2}$ and $b \in \croch{20, 30}$,
	\item Draw uniformly hyper-parameters for $\mukk$'s kernels : $\gammak = \{ v_{\gammak}, \ell_{\gammak} \}, \ \forall k \in \K$, where $v_{\gammak} \in \croch{1, e^3 }$ and $\ell_{\gammak} \in \croch{1, e^1}$, (or $\gamma_0 = \{ v_{\gamma_0}, \ell_{\gamma_0} \})$,
	\item Draw $\mukt \sim \mathcal{N} \paren{\mk(\Ut), \Ckt}, \forall k \in \K$,
	\item For all $i \in \I$, draw uniformly the hyper-parameters for individual kernels $\theta_i = \{ v_{\theta_i}, \ell_{\theta_i} \}$, where $v_{\theta_i} \in \croch{1, e^3 }$, $\ell_{\theta_i} \in \croch{1, e^1}$, and  $\sigma_i^2 \in \croch{0, 0.1}$, (or $\theta_0 = \{ v_{\theta_0}, \ell_{\theta_0} \}$ and $\sigma_0^2)$,
	\item Define $\pii = (\frac{1}{K}, \dots, \frac{1}{K})^{\intercal}$ and draw $\Zi \sim \mathcal{M}(1, \pii), \ \forall i \in \I$,
	\item For all $i \in \I$ and $\Zik = 1$, draw uniformly a random subset $\ti \subset \Ut$ of $N_i = 30$ timestamps, and draw $\yi \sim \mathcal{N} \paren{\muk(\ti), \Psii^{\ti}}$.
\end{enumerate}

This procedure offers data sets for both the individuals $\acc{\ti, \yi}_i$ and the underlying mean processes $\{ \Ut, \mukt \}_k$.
In the context of prediction, a new individual is generated according to the same scheme, although its first 20 data points are assumed to be observed while the remaining 10 serve as testing values. 
While it may be argued that this repartition 20-10 is somehow arbitrary, a more detailed analysis with changing numbers of observed points in \cite{LeroyMAGMAInferencePrediction2020} revealed a low effect on the global evaluation.
Unless otherwise stated, we fix the number of clusters to be $K^* = 3$ and the model assumption to be $\Hoo$ for generating the data.
Let us recall that we provided a variational-BIC formula in \Cref{prop:BIC} to select an appropriate number of clusters $K$ from data. 
Therefore, this measure is evaluated in following experiments and used for model selection purposes in the real-life application. 
\newline 

Besides, the adjusted rand index (ARI) \citep{HubertComparingpartitions1985} is used as a measure of adequacy for comparison between the groups obtained through the clustering procedure and the true clusters that generated the data.
More specifically, the ARI is defined by counting the proportions of matching pairs between groups, and a value of 1 represents a perfect correspondence.
One can note that the ARI still applies when it comes to evaluating clustering partitions with different numbers of clusters.
On the matter of prediction, the mean square error (MSE) between predicted means and the true values offers a measure of the average forecast performance.
Formally, we define the MSE in prediction on the 10 testing points for the new individual as:
\begin{equation*}
	\dfrac{1}{10} \sum\limits_{u = 21}^{30} \paren{ y_*^{pred} (t_*^u) - y_*^{true} (t_*^u) }^2.
\end{equation*}
Moreover, an additional measure accounting for the validity of uncertainty quantification is defined in \cite{LeroyMAGMAInferencePrediction2020} as the percentage of true data effectively lying within the $95 \%$ credible interval ($CI_{95}$), which is constructed from the predictive distribution.
We extend here this measure to the context of GPs mixture, where $CI_{95}$ is no longer available directly (as for any multi-modal distribution).
Namely, the weighted $CI_{95}$ coverage ($WCIC_{95}$) is defined to be: 
\begin{equation*}
	100 \times \dfrac{1}{10} \sum\limits_{u = 21}^{30} \sumk \taustk \ \mathds{1}_{ \{ y_*^{true}(t_*^u) \in \ CI_{95}^{k} \}},
\end{equation*}
\noindent where $CI_{95}^{k}$ represents the $CI_{95}$ computed for the $k$-th cluster-specific Gaussian predictive distribution \eqref{pred_clustspecific}.
In the case where $K=1$, i.e. a simple Gaussian instead of a GPs mixture, the $WCIC_{95}$ reduces to the previously evoked $CI_{95}$ coverage.
By averaging the weighted cluster-specific $CI_{95}^k$ coverage, we still obtain an adequate and comparable quantification of the uncertainty relevance for our predictions.
By definition, the value of this indicator should be as close as possible to $95\%$. 
Finally, the mean functions $\{\mk(\cdot) \}_k$ are set to be 0 in \algoclust, as usual for GPs, whereas the membership probabilities $\tauik$ are initialised thanks to a preliminary k-means algorithm.

\subsection{Illustration on Synthetic Examples}
\label{ch4sec:simu_illu_example}

\begin{figure*}
    \begin{center}
       \includegraphics[width=\textwidth]{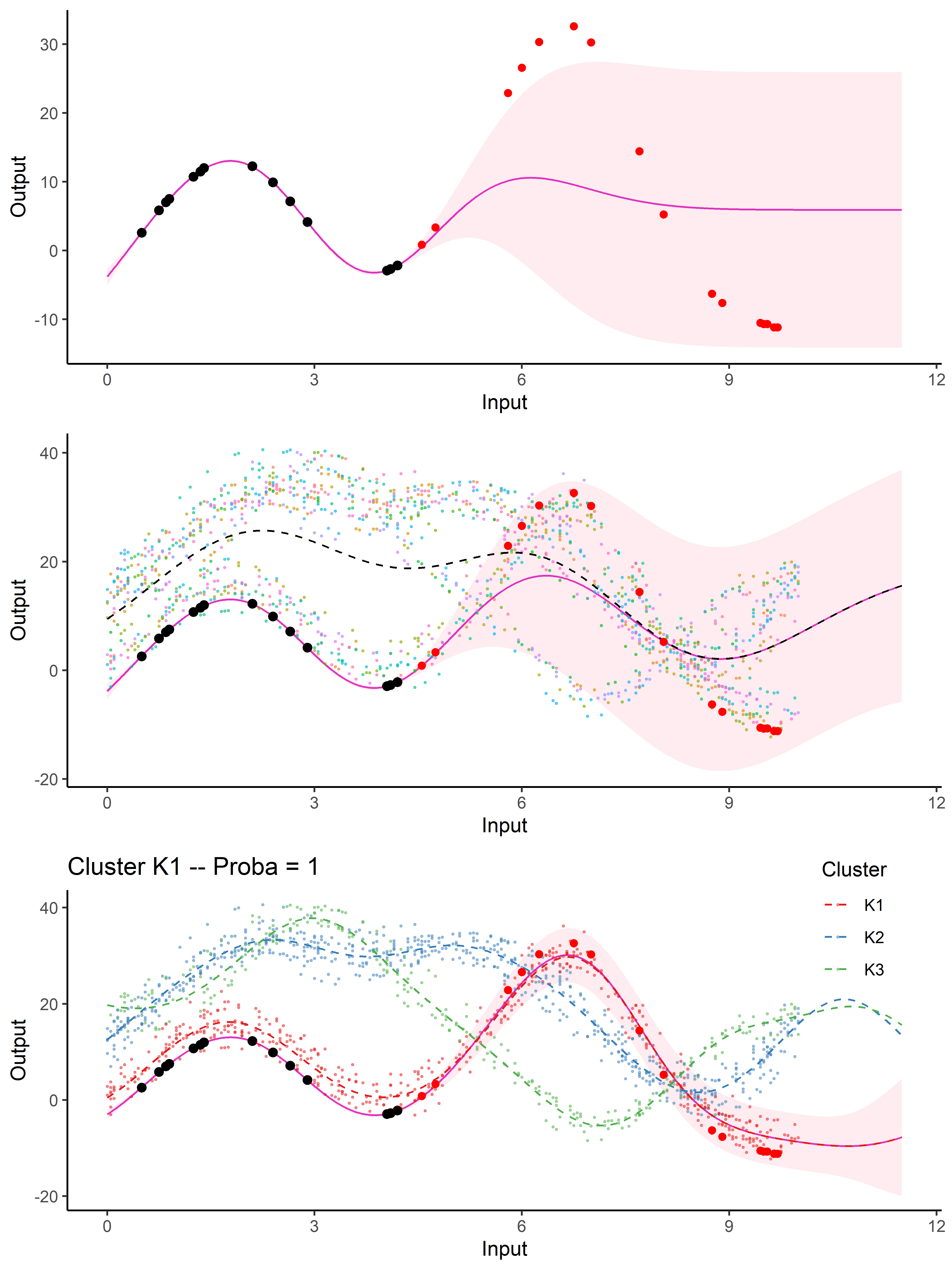}
       \caption{Prediction curves (blue) with associated 95\% credible intervals (grey) from GP regression (top), \algo (middle) and \algoclust (bottom). The dashed lines represent the mean parameters from the mean processes estimates. Observed data points are in black, testing data points are in red. Backward points are the observations from the training data set, coloured relatively to individuals (middle) or clusters (bottom).} 
		\label{illu_compare}  
    \end{center}
\end{figure*}
\begin{figure*}
    \begin{center}
       \includegraphics[width=\textwidth]{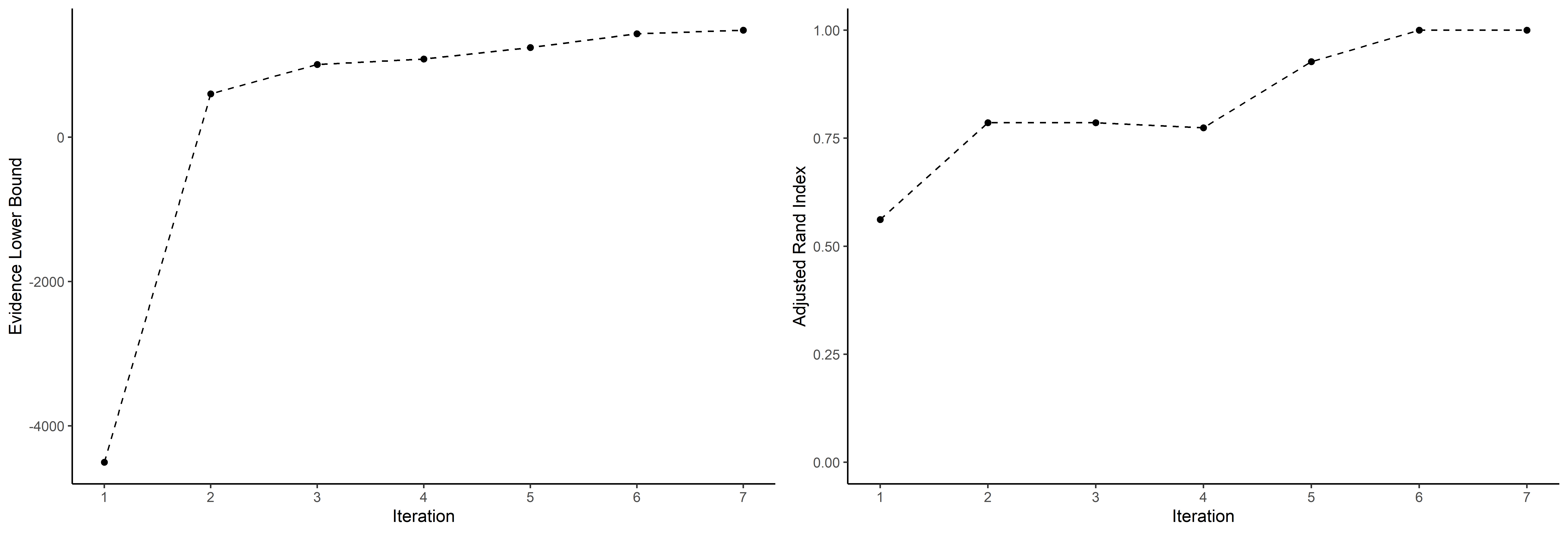}
       \caption{\textbf{Left:} Values of the evidence lower bound (ELBO) for successive iterations of the VEM algorithm during \algoclust training. \textbf{Right:} The corresponding ARI values between predicted and true clusters for the same iterations. In this illustrative example, the algorithm reached convergence after 7 iterations only.} 
		\label{elbo_ari}
    \end{center}
\end{figure*}

\Cref{illu_compare} provides a comparison on the same data set between a classical GP regression (top), the multi-task GP algorithm \algo (middle), and the multi-task GPs mixture approach \algoclust (bottom).
On each sub-graph, the plain blue line represents the mean parameter from the predictive distribution, and the grey shaded area covers the $CI_{95}$.
The dashed lines stand for the multi-task prior mean functions $\{ \mkhat(\cdot) \}_k$ resulting from the estimation of the mean processes. 
The points in black are the observations for the new individual $*$, whereas the red points constitute the true target values to forecast.
Moreover, the colourful background points depict the data of the training individuals, which we colour according to their true cluster in \algoclust displays (bottom).
For the sake of fairness, the prior mean for standard GP has been set as the average of training data points.
As expected, a simple GP regression provides an adequate fit close to the data points before quickly diving to the prior value 0 when lacking information.
Conversely, \algo takes advantage of its multi-task component to share knowledge across individuals by estimating a more relevant mean process. 
However, this unique mean process appears unable to account for the clear group structure, although adequately recovering the dispersion of the data.  
In the case of \algoclust , we display the cluster-specific prediction \eqref{pred_clustspecific} for the most probable group instead of the GPs mixture prediction, since $\max\limits_{k}(\taust) \approx 1$ in this example.
The model selection method based on maximum VBIC values correctly retrieved the true number of cluster $K = 3$.  
To illustrate the training procedure dynamics in this example, we provide on \Cref{elbo_ari} a tracking, for each iteration of the VEM algorithm until convergence, of the evidence lower bound (ELBO) and the corresponding ARI between predicted and true clusters. As the ELBO increases, notice that the ARI also improves to finally reach 1 at convergence, which means we fully retrieved the true clusters at the end of training. Although this simple example only depicts a single-run instance, it provides an accurate intuition on the general behaviour of \algoclust. In practice, the algorithm quickly improves the clustering structure and associated mean processes during the first two steps, and generally converges in a handful of iterations.
\newline

\begin{figure*}
    \begin{center}
       \includegraphics[width=\textwidth]{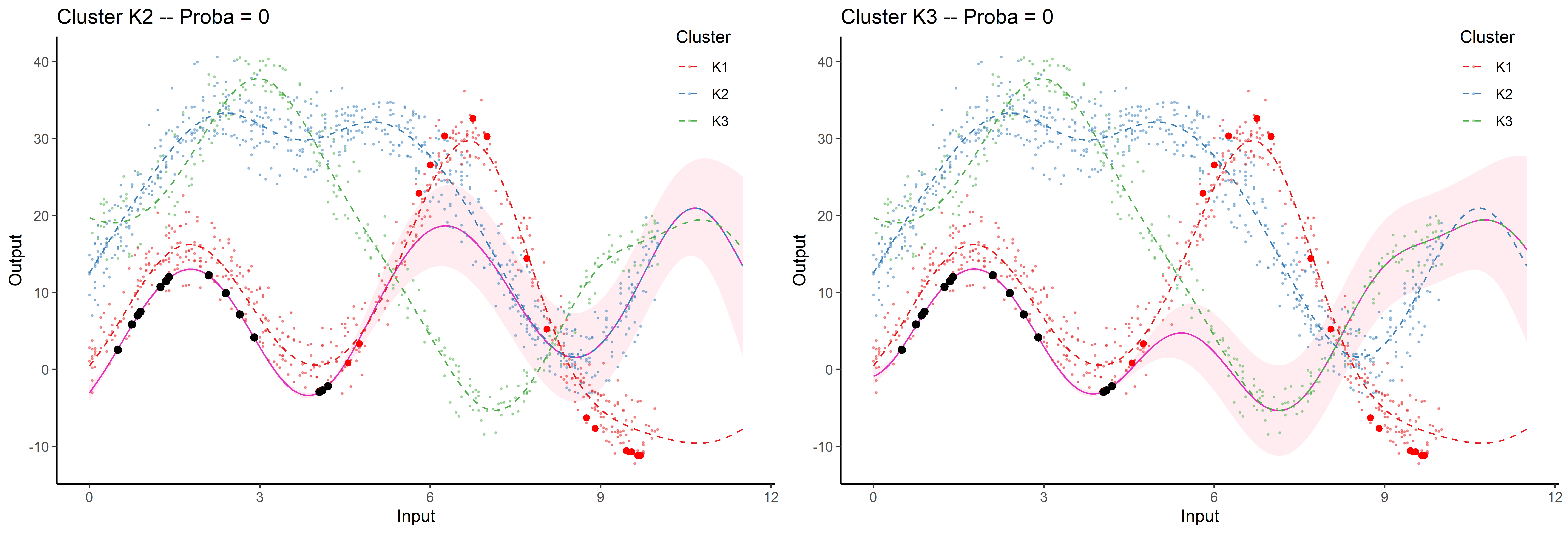}
       \caption{Cluster-specific prediction curves (blue) with associated 95\% credible intervals (grey) from \algoclust , for two unlikely clusters. The dashed lines represent the mean parameters from the mean processes estimates. Observed data points are in black, testing data points are in red. Backward points are the observations from the training data set, coloured by clusters.}
		\label{illu_2otherclusters}  
    \end{center}
\end{figure*}

\begin{figure*}
    \begin{center}
       \includegraphics[width=\textwidth]{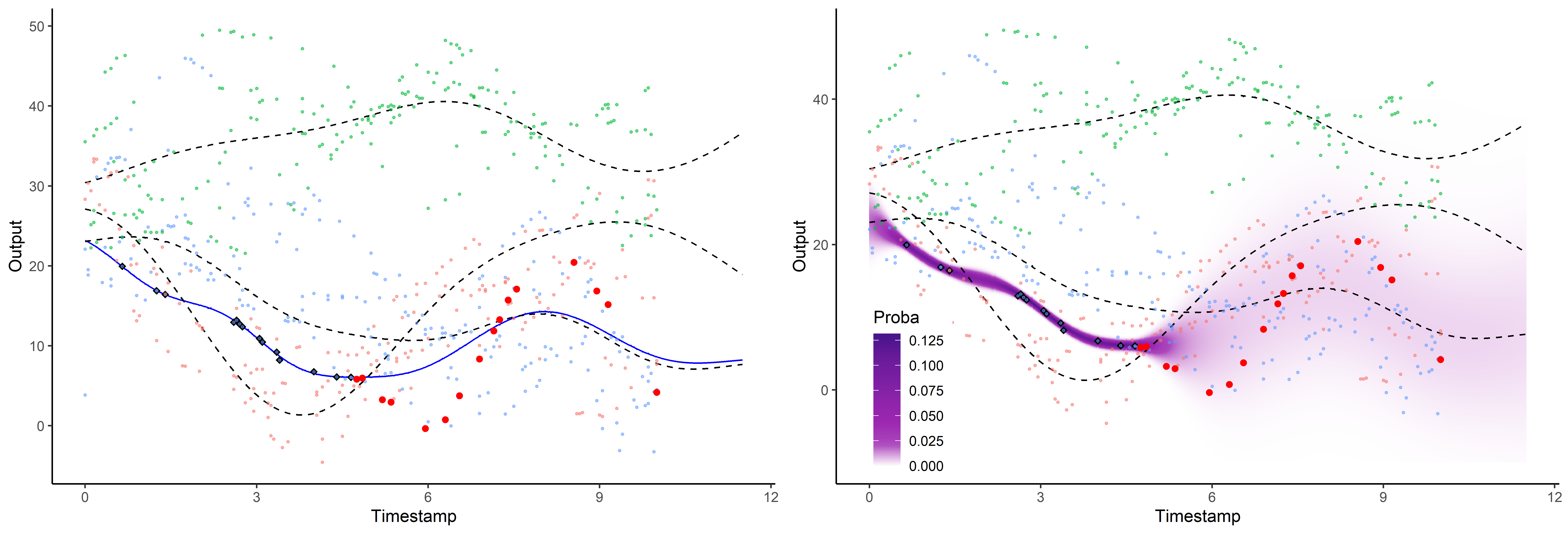}
       \caption{\textbf{Left:} GPs mixture mean prediction curve (blue) from \algoclust. \textbf{Right:} heatmap of probabilities for the GPs mixture predictive distribution from \algoclust . The dashed lines represent the mean parameters from the mean processes estimates. Observed data points are in black, testing data points are in red. Backward points are the observations from the training data set, coloured by clusters.}
		\label{illu_heatmap}
    \end{center}
\end{figure*}

Overall, this illustrative example highlights the benefit we can get from considering group-structured similarities between individuals in GP predictions. 
Notice that our method offers both a significant improvement in mean prediction and a narrowed uncertainty around this value. 
Additionally, we display on \Cref{illu_2otherclusters} the specific predictions according to the two remaining clusters (although associated with nearly 0 probabilities).
We remark that the predictions move towards the cluster specific mean processes as soon as the observations become too distant. 
In this idealised example, we displayed Gaussian predictive distributions for convenience though, in general, a Gaussian mixture might rarely be unimodal.
Therefore, we propose in \Cref{illu_heatmap} another example with a higher variance and overlapping groups, where the VBIC still provides the correct number of clusters. 
While the ARI between predicted and true clusters was equal to 1 (perfect match) in the previous example, it now decreases to 0.78.
Moreover, the vector of membership probabilities associated with the \Cref{illu_heatmap} for the predicted individual happens to be: $\taust = (0.95, 0.05, 0)$.
The left-hand graph provides an illustration of the predictive mean, acquired from the multi-task GPs mixture distribution described in \Cref{prop_finalpred}.
We may notice that this curve lies very close to one cluster's mean although not completely overlapping it, because of the $\taustk = 0.05$ probability for another cluster, which slightly pulls the prediction onto its own mean.
Besides, the right-hand graph of \Cref{illu_heatmap} proposes a representation of the multi-task GPs mixture distribution as a heatmap of probabilities for the location of our predictions. 
This way, we can display, even in this multi-modal context, a thorough visual quantification for both the dispersion of the predicted values and the confidence we may grant to each of them.
\newline

\begin{figure*}
    \begin{center}
       \includegraphics[width=\textwidth]{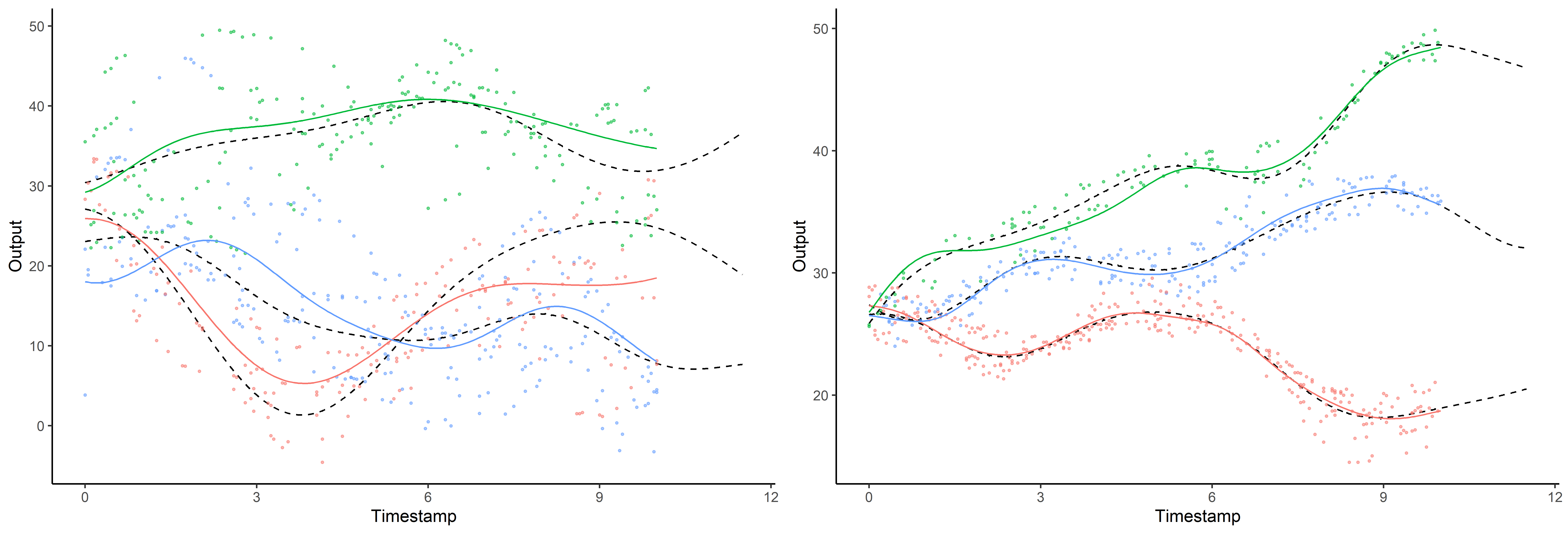}
       \caption{\textbf{Left:} fuzzy case. \textbf{Right:} well-separated case. Curves of the simulated underlying mean processes, coloured by clusters. The dashed lines represent the mean parameters from the mean processes estimates. Backward points are the observations from the training data set, coloured by clusters.} 
		\label{cluster_shape}  
    \end{center}
\end{figure*}

Finally, let us propose on \Cref{cluster_shape} an illustration of the capacity of \algoclust to retrieve the shape of the underlying mean processes, by plotting their estimations $\{ \mkhat (\cdot)\}_k$ (dotted lines) along with the true curves (plain coloured lines) generated by the simulation scheme. 
The ability to perform this task generally depends on the structure of the data as well as on the initialisation, although we may observe satisfactory results both on the previous fuzzy example (left) and on a well-separated case (right). 
It should be noted that the mean processes' estimations also come with uncertainty quantification, albeit not displayed on \Cref{cluster_shape} for the sake of clarity. 

\subsection{Clustering Performance}

\begin{figure}
    \begin{center}
       \includegraphics[width=.7\textwidth]{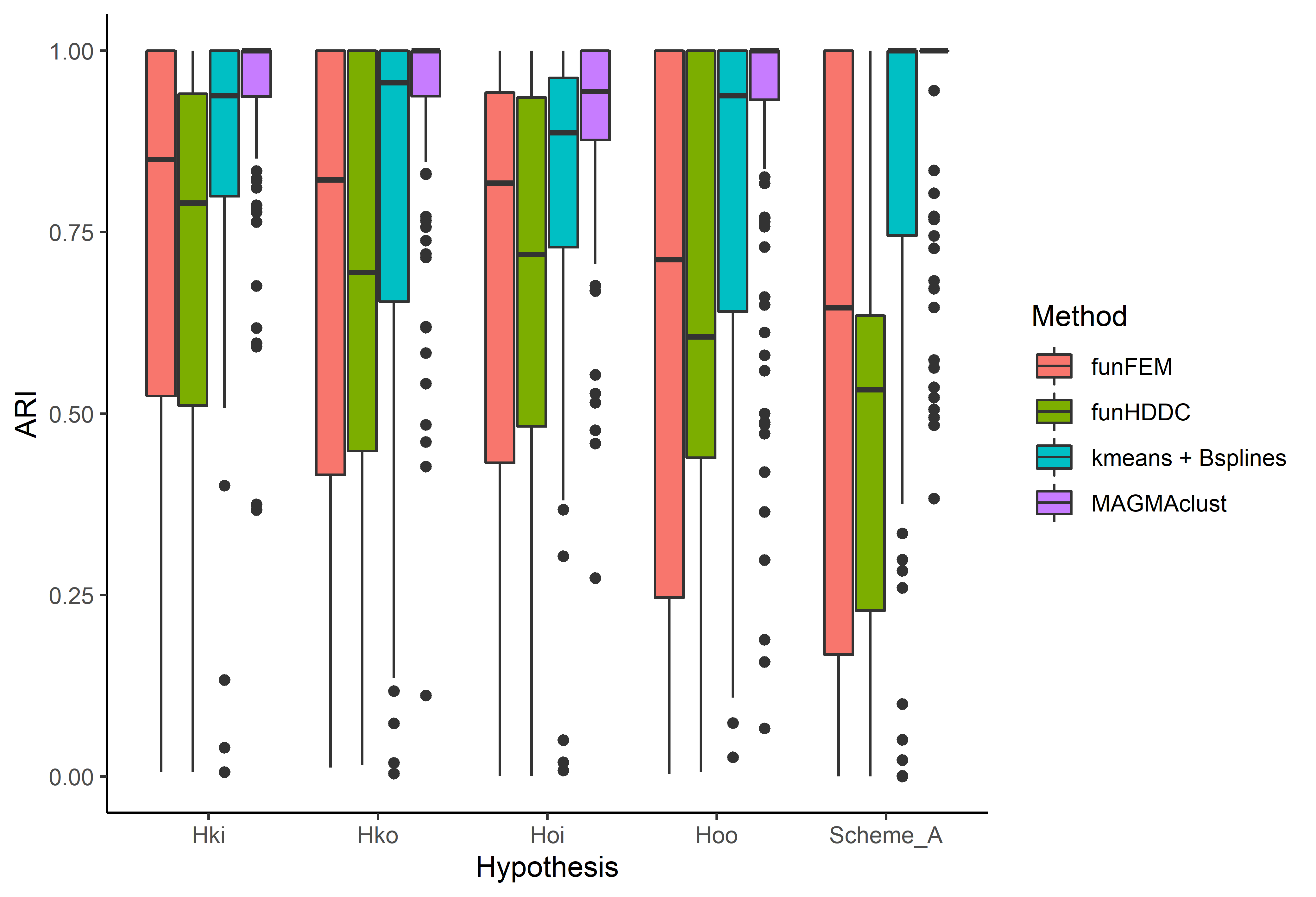}
       \caption{Adjusted rand index (ARI) values between the true clusters and the partitions estimated by kmeans, funHDDC, funFEM, and \algoclust . The value of K is set to the true number of clusters for all methods. The ARI is computed on 100 data sets for each generating model's assumption $\Hki, \Hko, \Hoi$, $\Hoo$, and \emph{Scheme A}.}
		\label{compare_clust}  
    \end{center}
\end{figure}

\begin{figure}
    \begin{center}
       \includegraphics[width=0.7\textwidth]{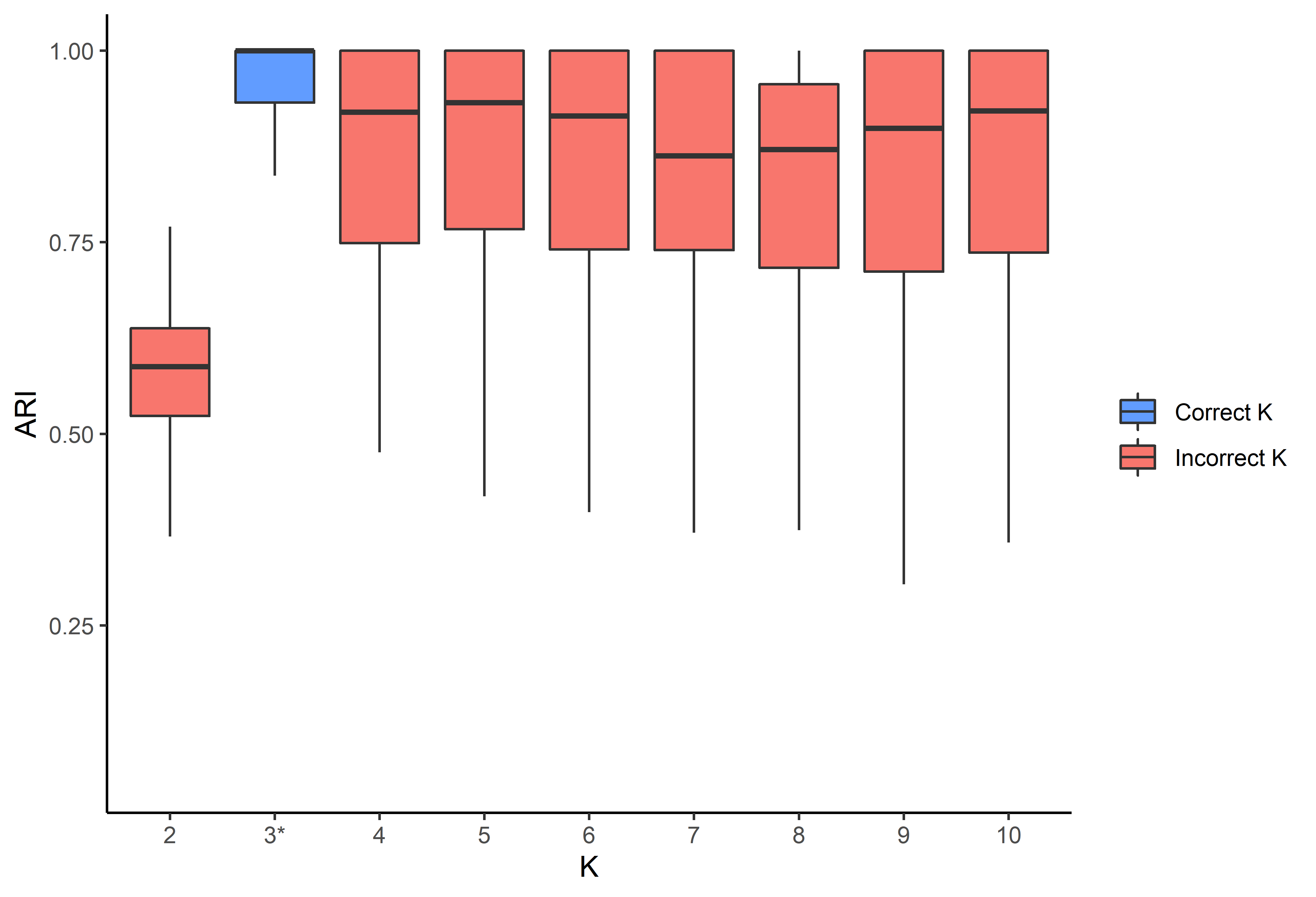}
       \caption{Adjusted Rand Index values between the true clusters and the partitions estimated by \algoclust with respect to the values of $K$ used as setting. The ARI is computed on the same 100 data sets for each value of K. ($3^*$: the true number of clusters for all data sets)} 
		\label{clust_diffk}  
    \end{center}
\end{figure}

Generally, many curve clustering methods struggle to handle irregularly observed data directly.
Therefore, for the sake of fairness and to avoid introducing too many smoothing biases in alternative methods, the data sets used in the following are sampled on regular grids, although \algoclust can deal with irregular measurements by construction (see also \cite{LeroyMAGMAInferencePrediction2020} for an empirical study).
The competing algorithms are the B-splines expansion associated with a kmeans algorithm proposed in \cite{AbrahamUnsupervisedCurveClustering2003}, funHDDC \citep{Bouveyron2011}, and funFEM \citep{bouveyron2015discriminative}. 
The two latter methods were introduced to handle curve clustering problems, by taking advantage of a functional latent mixture modelling, and demonstrated their ability in several applications \citep{SchmutzClusteringmultivariatefunctional2018}. 
A naive multivariate kmeans is used as initialisation for funHDDC, funFEM, and \algoclust.
We propose on \Cref{compare_clust} an evaluation of each algorithm in terms of ARI over 100 data sets, simulated from various schemes. 
First, the procedure detailed in \Cref{sec:exp} is applied for each of the 4 different hypotheses ($\Hki, \Hko, \Hoi, \Hoo$) to generate data in accordance with our modelling assumptions. 
Additionally, we propose an alternative simulation scheme, inspired by \cite{bouveyron2015discriminative}, to compare performances on data sets that are not tailored for our method. We name this procedure \emph{Scheme A}, and each of the 100 data sets is made of 50 curves, generated randomly, allocated into 4 clusters, and observed at 30 common time points such that:
\begin{itemize}
	\item Cluster 1: $y(t) = U + 0.5 \times (1 - U ) \times  (2.5 - \abs{t - 2.5})^{+} + \epsilon, \ \ \ \ t \in [0, 10],$
	\item Cluster 2: $y(t) = U + 0.5 \times (1 - U ) \times  (2.5 - \abs{t - 7.5})^{+} + \epsilon, \ \ \ \ t \in [0, 10],$
	\item Cluster 3: $y(t) = U + (1 - U ) \times  (2.5 - \abs{t - 2.5})^{+} + \epsilon, \ \ \ \ t \in [0, 10],$
	\item Cluster 4: $y(t) = U + (1 - U ) \times  (2.5 - \abs{t - 7.5})^{+} + \epsilon, \ \ \ \ t \in [0, 10],$ 
\end{itemize} 
where $U \sim \mathcal{U}([0,1])$ and $\epsilon \sim \mathcal{N}(0, 0.05)$.

In all situations, \algoclust seems to outperform the alternatives .
In particular, our approach provides ARI values consistently close to 1 and a lower variance in all contexts. 
Furthermore, while performances of the other methods are expected to deteriorate because of additional smoothing procedures in the case of irregular grids, \algoclust would run the same without any change.
\newline

On another aspect, \Cref{clust_diffk} provides some insights into the robustness of \algoclust to an incorrect setting of $K$, the number of clusters.
For 100 data sets with a true value $K^* = 3$, the ARI has been computed between the true partitions and the ones estimated by \algoclust initialised with different settings $K =  2, \dots, 10$. 
Except for $K=2$ where the low number of clusters prevents from getting enough matching pairs by definition, we may notice relatively unaffected performances as $K$ increases.
Despite a non-negligible variance in results, the partitions remain consistent overall, and the clustering performances of \algoclust seem pretty robust to misspecification of $K$.

\subsection{Model Selection Performance}

\begin{table}[htbp]
\begin{center}
\begin{tabular}{c|cccccc|cccccc|}
\cline{2-13}
\multicolumn{1}{l|}{}         & \multicolumn{12}{c|}{Selected K}                                                                                                                    \\
\multicolumn{1}{l|}{}         & \multicolumn{6}{c}{M = 50}                                               & \multicolumn{6}{c|}{M = 100}                                             \\ \hline
\multicolumn{1}{|c|}{True K*} & 1            & 2           & 3           & 4           & 5           & 6 & 1            & 2           & 3           & 4           & 5           & 6 \\ \hline
\multicolumn{1}{|c|}{1}       & \textbf{100} & 0           & 0           & 0           & 0           & 0 & \textbf{100} & 0           & 0           & 0           & 0           & 0 \\
\multicolumn{1}{|c|}{2}       & 10           & \textbf{90} & 0           & 0           & 0           & 0 & 2            & \textbf{96} & 2           & 0           & 0           & 0 \\
\multicolumn{1}{|c|}{3}       & 0            & 22          & \textbf{74} & 2           & 2           & 0 & 2            & 14          & \textbf{84} & 0           & 0           & 0 \\
\multicolumn{1}{|c|}{4}       & 4            & 10          & 16          & \textbf{58} & 10          & 2 & 2            & 10          & 8           & \textbf{80} & 0           & 0 \\
\multicolumn{1}{|c|}{5}       & 0            & 10          & 16          & 20          & \textbf{52} & 2 & 0            & 0           & 14          & 18          & \textbf{68} & 0 \\ \hline
\end{tabular}
\end{center}
\caption{Percentage of data sets for which the true number of cluster is $K^*$, and the number of cluster selected by the VBIC is $K$. A total of 50 data sets were simulated for different values $K^* = 1,\dots, 5$, and \algoclust was tested on each with varying settings $K = 1,\dots, 6$, to retain the configuration that reaches the highest VBIC value. The data sets are composed of $M = 50$ (left) or $M=100$ (right) individuals with $N = 30$ common timestamps, under the hypothesis $\Hoo$.}
\label{table_BIC}
\end{table}

To remain on the matter of clusters' number, the model selection abilities of the proposed VBIC (\Cref{prop:BIC}) maximisation procedure are assessed on simulated data.
For different true numbers of clusters, 50 data sets were simulated according to the previous scheme, and \algoclust was run successively on each with different settings of K.
The setting that reaches the highest value of VBIC is selected as the best model. 
The percentages of selection for each true $K^*$ and the corresponding values $K$ retained through VBIC are reported in \Cref{table_BIC}.
The procedure seems to adequately select the number of clusters in most context, with results that deteriorate as $K$ grows and a tendency to over-penalise, which is a classical behaviour with BIC in practice \citep{WeakliemCritiqueBayesianInformation2016}. 
As the typical BIC formula relies on asymptotic approximations, we also ran the simulation for different numbers of individuals $M = 50$ and $M = 100$.
It may be noticed that the VBIC performs better as $M$ increases, as expected.
Such a property appears reassuring since the following real-data applications involves data sets gathering around $M = 10^3$ individuals.

\subsection{Prediction Performance}

\begin{table}[htbp]
\begin{center}
\begin{tabular}{|c|cc|cc|}
\hline
K           & MSE                  & $WCIC_{95}$                 & Training time & Prediction time \\ \hline
2           & 7.7 (18.4)          & 92 (20.3)           & 70.4 (25)    & 0.4 (0.1)     \\
\textbf{3*} & 3.7 (8.1)          & \textbf{95 (13.2)} 		 & 97.7 (33.2)   & 0.5 (0.1)     \\
4           & 3.2 (5.3)          & 94.9 (13.6) 		 & 116.5 (47.3)   & 0.6 (0.2)     \\
5           & 3.2 (5.6) 		   & 94.4 (14.3)          & 133 (40.8)   & 0.6 (0.2)     \\
6           & \textbf{3.1 (5.4)}          & 94.4 (13.6)          & 153.3 (42)   & 0.8 (0.3)     \\
7           & 4 (9)              & 93.6 (15.4)          & 173.7 (45.1)   & 1 (0.4)     \\
8           & 4.7 (13)           & 93.8 (16)            & 191.3 (44.7)   & 1 (0.3)     \\
9           & 4.1 (9.5)          & 94 (14.6)            & 211.6 (52)    & 0.8 (0.4)     \\
10          & 4.5 (14.8)         & 94.4 (14.4)          & 235 (52.7)    & 1.8 (1.4)     \\ \hline
\end{tabular}
\end{center}
\caption{Average (sd) values of MSE, $WCIC_{95}$, training and prediction times (in secs) on 100 runs for different numbers of clusters as setting for \algoclust. ($3^*$ : the true number of clusters for all data sets)}
\label{table_diffk_pred}
\end{table}

Another piece of evidence for the robustness to a wrong selection of K is highlighted by \Cref{table_diffk_pred} in the context of forecasting. 
The predictive aspect of \algoclust remains the main purpose of the method and its performances of this task partly rely on the adequate clustering of the individuals. 
It may be noticed on \Cref{table_diffk_pred} that both MSE and $WCIC_{95}$ regularly but slowly deteriorate as we move away from the true value of K.
However, the performances remain of the same order, and we may still be confident about the predictions obtained through a misspecified running of \algoclust .
In particular, the values of MSE happen to be even better when setting $K = 4, \dots, 6$ (we recall that the same 100 data sets are used in all cases, which can thus be readily compared). 
Besides, the right-hand part of the table provides indications on the relative time (in seconds) that is needed to train the model for one data set and to make predictions. 
As expected, both training and prediction times increase roughly linearly with the values of $K$, which seems consistent with the complexities exposed in \Cref{sec:complexity}.
This property is a consequence of the extra mean processes and hyper-parameters that need to be estimated as $K$ grows.
Nonetheless, the influence of $K$ is lesser on the prediction time, which yet remains negligible, even when computing many group-specific predictions.
\newline

To pursue the matter of prediction, we provide on \Cref{table_compare_pred} the comparison of forecasting performances between our method and several state-of-the art alternatives. 
We use the classical (i.e. single-task) GP regression as benchmark, as we would expect all the competing multi-tasks methods to perform better, considering the additional information that can be shared between individuals.
The \algo algorithm \citep{LeroyMAGMAInferencePrediction2020} (which is equivalent to \algoclust in the specific case where $K = 1$) is also evaluated to measure the accuracy gain we can achieve thanks to the additional clustering on the simulated group-structured data sets.
As detailed in \Cref{sec:intro}, the existing multi-task approaches in the GP literature generally consider specific kernel structures accounting for explicit correlations both between inputs and outputs \citep{AlvarezKernelsVectorValuedFunctions2012a}.
The Multi-Output Gaussian Process Tool Kit (MOGPTK) \citep{dewolffMOGPTKMultioutputGaussian2021} is a Python package that implements the main multi-output covariance kernels from literature in a unified framework. 
We relied on this package to run experiments, first applying the proposed combination (SM LCM) of the spectral mixture (SM) \citep{wilson13} and the linear model of coregionalisation (LMC) \citep{goovaerts1997geostatistics}, which is the historical and very general formulation for multi-output kernels. 
Additionally, we made use of the more recent multi-Output spectral mixture (MOSM) algorithm  \citep{parra2017spectral}, which is defined as the default method in the package.
For the sake of fair comparison, the L-BFGS-B algorithm \citep{NocedalUpdatingquasiNewtonmatrices1980,MoralesRemarkalgorithmLBFGSB2011} is the optimisation procedure used in all competing methods.  
\newline

Regarding both mean prediction and uncertainty quantification, our approach outperforms the alternatives. 
In terms of MSE, \algoclust takes advantage of its multiple mean processes to results in an order of magnitude enhancement compared to the best competitors (\algo and SM LMC). 
As expected, the simple GP regression performs rather poorly and surprisingly MOSM results are even worse, as in practice, its current implementation seems to reach pathological cases during training most of the time.
Additionally, the empirical quantification of uncertainty of \algoclust appears very convincing since there are on average exactly $95\%$ of the observations lying within the weighted $CI_{95}$, as expected. 
\newline

In accordance with the theoretical complexity, the increase in training times displayed in \Cref{table_compare_pred} remains roughly proportional to the value of $K$ (we recall that \algoclust assumes $K=3$ here, compared to \algo which is $K=1$).
In contrast, the multi-output GPs methods (SM LMC and MOSM) multiply tenfold both training and prediction times even in these reasonable experiments (50 individuals and 30 timestamps). 
This cost comes from the cubic complexity in the number of tasks (due to size-augmented covariance matrices) that results in a massive computational burden as $M$ increases.
As a consequence, in the following real data applications, where thousands of individuals are considered, these methods would merely be unusable in practice.

\begin{table}
\begin{center}
\begin{tabular}{c|cc|cc|}
\cline{2-5}
                                                & MSE         & $WCIC_{95}$        & Training time & Prediction time \\ \hline
\multicolumn{1}{|c|}{GP}                        & 138 (174)   & 78.4 (31.1)  & 0 (0)         & 0.6 (0.1)     \\
\multicolumn{1}{|c|}{SM LMC}                        & 29.9 (95.9)   & 97.2 (7)  & 1172.6 (300)         & 5.6 (0.6)     \\
\multicolumn{1}{|c|}{MOSM}                        & 417 (646)   &  57.8 (45.2)  & 1148 (60)         & 7.5 (0.5)     \\
\multicolumn{1}{|c|}{\algo}    				    & 31.7 (45) & 84.4 (27.9) & 61.1 (25.7)   & 0.5 (0.2)     \\
\multicolumn{1}{|c|}{\algoclust} & \textbf{3.7 (8.1)} & \textbf{95 (13.2)}  & 132 (55.6)   & 0.6 (0.2)     \\ \hline
\end{tabular}
\end{center}
\caption{Average (sd) values of MSE, $WCIC_{95}$, training and prediction times (in secs) for GP, SM LMC, MOSM, \algo and \algoclust over 100 simulated testing sets.}
\label{table_compare_pred}
\end{table}

\subsection{Application to real data sets}
To evaluate the efficiency of our approach in real-life applications, we introduce 3 distinct data sets corresponding to various contexts: sports performances, weight follow-up during childhood, and missing data reconstruction in $CO_2$ emissions.
The common aspect in all these problems lies in the presence of time series collected from multiple sources, possibly with irregular measurements, for which we expect to provide accurate forecasts by taking advantage of shared information and clustered structures in the data.
For all experiments, the individuals (or countries for $CO_2$) are split into training and testing sets (in proportions $60\% - 40\%$). 
In the absence of expert knowledge, the prior mean functions $\{ m_k (\cdot) \}_k$ are set to be constant equal to 0. 
The hypothesis $\Hoo$ is specified along with random initialisations for the hyper-parameters.
The hyper-parameters, the mean processes and the cluster's membership probabilities are learnt on the training data set. 
Then, the data points of each testing individual are split for evaluation purposes between observed (the first 60\%) and testing values (the remaining 40\%).
Therefore, each new process $y_*(\cdot)$ associated with a test individual is assumed to be partially observed, and its testing values are used to compute MSE and $WCIC_{95}$ for the predictions. 
As measuring clustering performances directly for real-life applications is a vain effort  \citep{kleinberg2002impossibility}, the results are provided for several values of $K$, from $K = 1$ (i.e. \algo) to $K  = 6$, to evaluate the effect of clustered-data structures on predictions. 
In all the following experiments, predictive performances tend to reach a plateau as we increase the number of clusters and no substantial improvement is noticeable for $K \geq 7$.

\subsubsection*{Talent identification in competitive swimming}

As previously presented in \Cref{sec:motivation}, the 100m freestyle swimming data sets initially proposed in \cite{LeroyFunctionalDataAnalysis2018} and \cite{LeroyMAGMAInferencePrediction2020} is analysed below in the new light of \algoclust .
The data sets contain results from 1731 women and 7876 men, members of the French swimming federation, each of them compiling an average of 22.2 data points (min = 15, max = 61) and 12 data points (min = 5, max = 57) respectively.
In the following, age of the $i$-th swimmer is considered as the input variable (timestamp $t$) and the performance (in seconds) on a 100m freestyle as the output ($y_i(t)$).
The analysis focuses on the youth period, from 10 to 20 years, where the progression is the most noticeable.
Let us recall that we aim at modelling a curve of progression from competition results for each individual in order to forecast their future performances.
We expect \algoclust to provide relevant predictions by taking advantage of both its multi-task feature and the clustered structure of data, previously highlighted in \cite{LeroyFunctionalDataAnalysis2018}.
\newline

\begin{table}
\begin{center}
\begin{tabular}{c|cc|cc|}
\cline{2-5}
 & \multicolumn{2}{c|}{Women} & \multicolumn{2}{c|}{Men} \\  
 & Mean & $WCIC_{95}$ & Mean & $WCIC_{95}$ \\ \hline
\multicolumn{1}{|c|}{\algo} & 4.9 (6.6) & 94.3 (14.2) & 3 (3.3) & 96.9 (9.3) \\
\multicolumn{1}{|c|}{\algoclust - K = 2} & 4.5 (6.2) & 94.3 (14.1) & 2.8 (3.2) & 96.2 (10.1) \\
\multicolumn{1}{|c|}{\algoclust - K = 3} & 4.3 (6) & \textbf{94.4 (14.1)} & 2.8 (3.1) & \textbf{96.1 (10.4)} \\
\multicolumn{1}{|c|}{\algoclust - K = 4} & 4.2 (5.9) & \textbf{94.4 (14.1)} & 2.8 (3) & 96.2 (10) \\
\multicolumn{1}{|c|}{\algoclust - K = 5} & 4.2 (5.9) & \textbf{94.4 (14.1)} & \textbf{2.7 (3)} & \textbf{96.1 (10.1)} \\
\multicolumn{1}{|c|}{\algoclust - K = 6} & \textbf{4.1 (5.7)} & 94.3 (14.1) & 2.8 (3) & 96.3 (9.8) \\ \hline
\end{tabular}
\end{center}
\caption{Average (sd) values of MSE and $WCIC_{95}$ for \algoclust with $K = 1, \dots, 6$ on the french swimmer
testing data sets.}
\label{table_real_data}
\end{table}

\begin{figure*}
    \begin{center}
       \includegraphics[width=\textwidth]{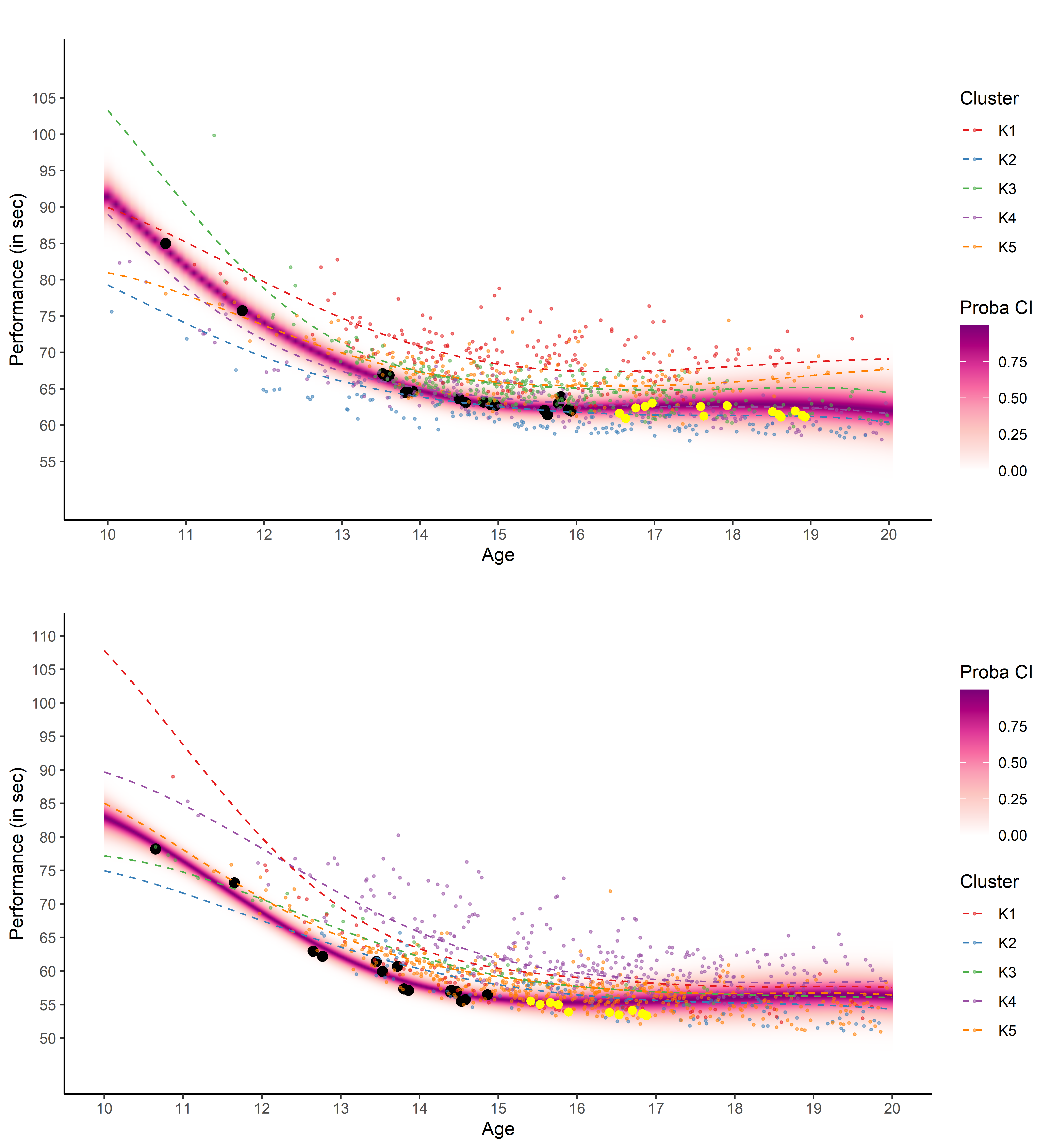}
       \caption{\textbf{Top:} women data set. \textbf{Bottom:} men data set. Heatmap of probabilities obtained through the GPs mixture predictive distribution of \algoclust with $K = 5$ for a random illustrative swimmer. The dashed lines represent the mean parameters from the mean processes estimates. Observed data points are in black, testing data points are in yellow. Backward points are a sample of observations from the training data set, coloured according to their most probable cluster.} 
		\label{swimming_data}
    \end{center}
\end{figure*}
As exhibited by \Cref{table_real_data}, \algoclust offers excellent performances on both data sets and slightly improves \algo predictions, as we increase the number of clusters. 
Values of both MSE and $WCIC_{95}$ appear satisfactory in all cases, and cluster-specific predictions provide additional accuracy though one may fairly argue that the gain remains moderate overall. 
One of the explaining reasons is highlighted in \Cref{swimming_data}, where we displayed illustrative predictions for a random man and woman when $K = 5$.
Although we can notice clear distinctions between the different mean curves at young ages, these differences tend to decrease afterwards, as adults' performances lie in narrow intervals, especially in regards to the overall signal-on-noise ratio.
Nevertheless, \algoclust provides several additional insights into this problem compared to \algo .
\newline 

First, the clusters offer interesting results to assess the profile of young swimmers and to determine the individuals to whom they most resemble. 
In particular, it is also possible to differentiate future evolutions associated with each cluster, along with their probabilities to occur (we do not display all the cluster-specific predictions here for the sake of concision).
On the other hand, our method leads to tighter predictive distributions in terms of uncertainty.
Compared to \algo which uses all training data identically, \algoclust somehow discards the superfluous information, through the weights $\taustk$, to only retain the most relevant cluster-specific mean processes. 
\newline 


\subsubsection*{Follow-up of children's weight in Singapore}

\begin{table}
\begin{center}
\begin{tabular}{c|cc|}
\cline{2-3}
 & MSE & $WCIC_{95}$ \\ \hline
\multicolumn{1}{|l|}{\algo} & 5.35 (9.48) & 93.3 (16.7) \\
\multicolumn{1}{|l|}{\algoclust - K = 2} & 4.94 (8.98) & \textbf{95.1 (15.2)} \\
\multicolumn{1}{|l|}{\algoclust - K = 3} & 5.01 (9.67) & 94.7 (16) \\
\multicolumn{1}{|l|}{\algoclust - K = 4} & 4.90 (9) & 94.8 (16.1) \\
\multicolumn{1}{|l|}{\algoclust - K = 5} & 4.95 (9.18) & 94.7 (16.1) \\
\multicolumn{1}{|l|}{\algoclust - K = 6} & \textbf{4.85 (9.8)} & \textbf{94.9 (16.1)} \\ \hline
\end{tabular}
\end{center}
\caption{Average (sd) values of MSE and $WCIC_{95}$ for \algoclust with $K = 1, \dots, 6$ on the children's weight testing data set.}
\label{table_weight_data}
\end{table}

In contrast with the previous application, we now study individual time series that are almost similar at young ages before diverging while growing older.
This data set (collected through the GUSTO program, see \url{https://www.gusto.sg/}) corresponds to a weight follow-up of 342 children from Singapore at 11 timestamps between birth and 72 months.
In this experiment, the goal is to predict the weight of a child at $\{24, 36 , 48, 60, 72\}$ months, using its observed weight at $\{0, 3, 6, 9 ,12, 18 \}$ months and data from the training individuals. 
Since the weight differences between toddlers are shallow until 18 months, providing accurate long-term forecasts, while clear morphology differences emerge, seems particularly challenging at first sight.
However, \algoclust still achieves impressive performances in this context as highlighted by \Cref{table_weight_data}.
Once again, the accuracy of predictions seems to slightly improve as we increase the number of clusters.
In this application, recovering an adequate cluster for a child is essential to anticipate which future weight pattern is the most likely, and assuming that more clusters exist appears to help in identifying specific patterns more precisely. 
\newline

\begin{figure*}
    \begin{center}
       \includegraphics[width=\textwidth]{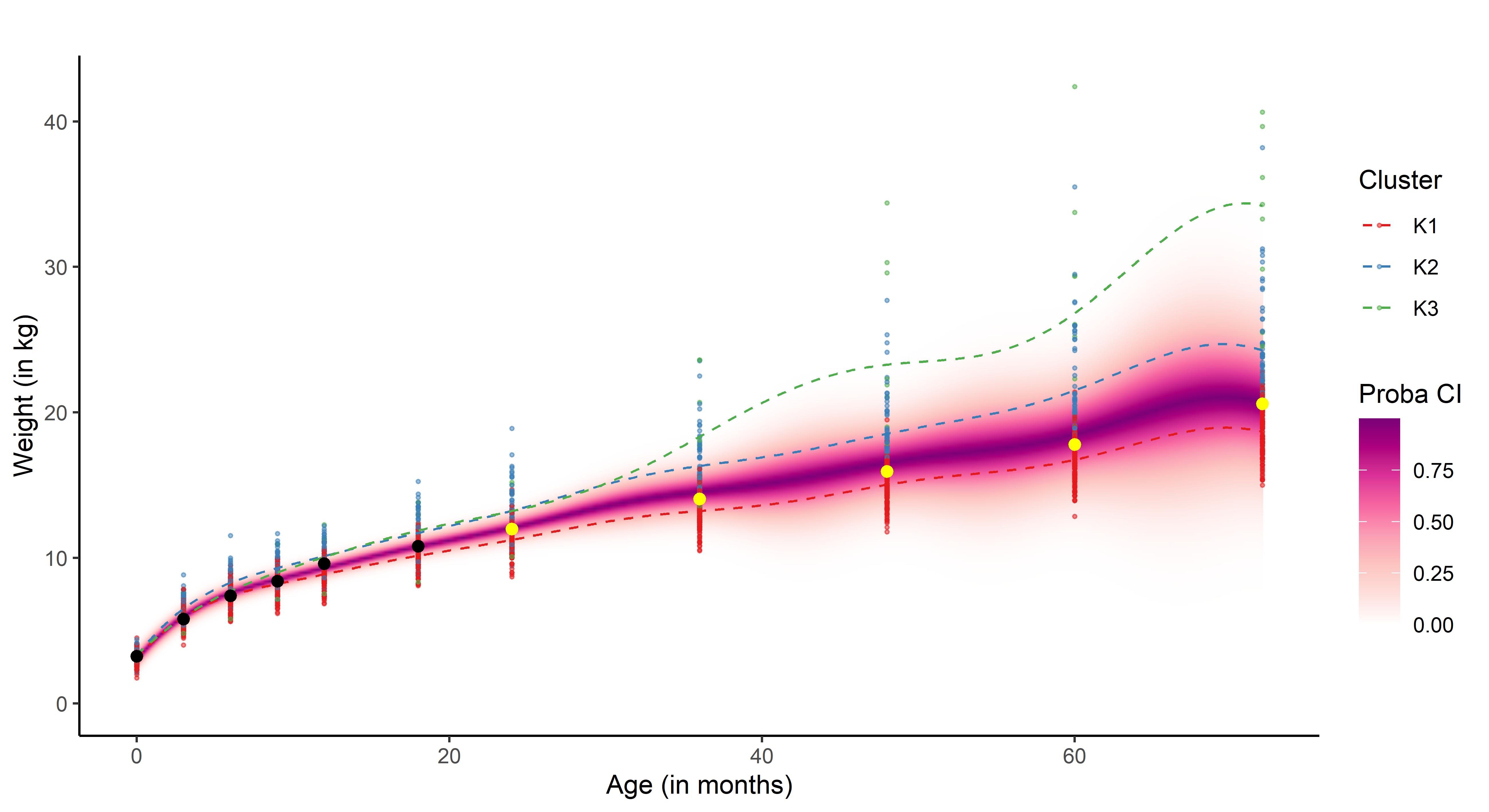}
       \caption{Heatmap of probabilities obtained through the GPs mixture predictive distribution of \algoclust with $K = 5$ for a random illustrative swimmer. The dashed lines represent the mean parameters from the mean processes estimates. Observed data points are in black, testing data points are in yellow. Backward points are observations from the training data set, coloured according to their most probable cluster.}
		\label{fig:real_weight}
    \end{center}
\end{figure*}

This being said, each new cluster increases the overall complexity of the model and whether it is worth adding one cluster regarding the relative gain in accuracy remains a practitioner's choice. 
For instance, the VBIC measure we proposed in \Cref{sec:model_selection} indicates $K = 3$ as the optimal number of clusters for this data set (keep in mind that this criterion maximises a penalised ELBO and tells us nothing regarding the predicting abilities). 
Using this value for $K$ allows us to display on \Cref{fig:real_weight} the behaviour of \algoclust predictions for a random testing child.
Notice that the prediction remains nicely accurate all along even though the testing points (in yellow) are not close to any of the mean processes.
This particularity comes from the learning of cluster weights $\taustk$ at young ages, where the algorithm identifies that this child belonged in nearly equal proportions to the clusters K1 (in red) and K2 (in blue).
Therefore, the multi-task GPs mixture posterior distribution defined in \Cref{prop_finalpred} defines a weighted trade-off between the two mean processes that remarkably predicts the true weight values of the child for almost 5 years.
Although this nice behaviour on a single example does prevent pathological cases to occur in general, we know from \Cref{table_weight_data} that those remain rare in practice, and that the computed credible intervals encompass true data with the correct degree of uncertainty. 

\subsubsection*{Reconstructing missing data in $CO_2$ emissions}

For this last application, we propose to use \algoclust to tackle a different kind of problem, namely, missing data reconstruction. 
Contrarily to the previous data sets focused on forecasting, the present collection of time series consists in historical measurements of $CO_2$ emissions per capita for each country from 1750 to 2020 (freely available at \url{https://github.com/owid/co2-data}).
Naturally, most countries in the world have not collected such data regularly and many annual observations are missing, especially as we move back in the past (for instance, only Canada and the United Kingdom reported values before 1800).
However, our method seems particularly well-suited to recover probable distributions of historical $CO_2$ emissions by exploiting similarities between countries and transferring knowledge from the densely observed time series to those that remain sparse. 
\newline 
\begin{figure*}
    \begin{center}
       \includegraphics[width=\textwidth]{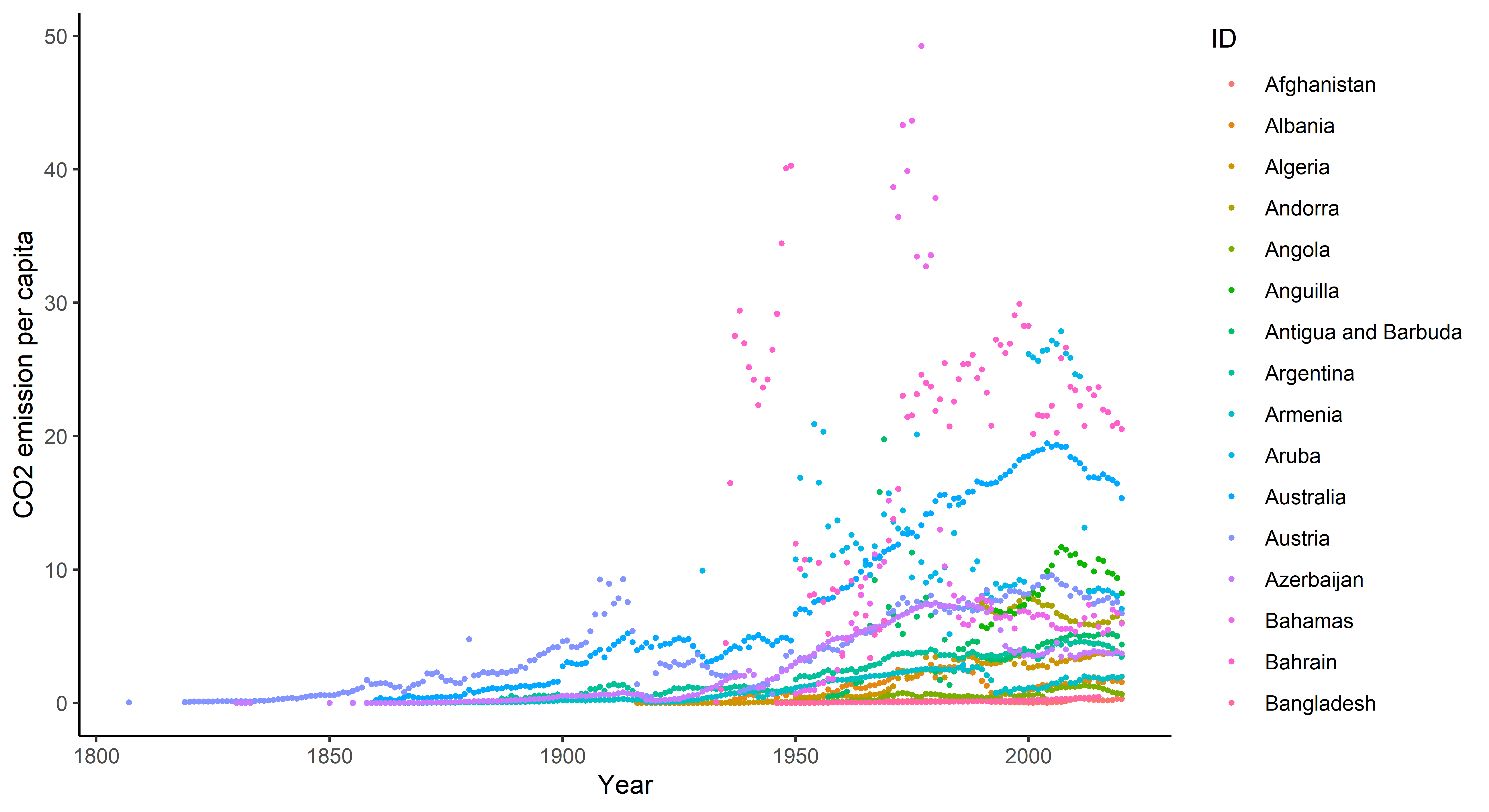}
       \caption{Time series representing the evolution of $CO_2$ emissions per capita for an illustrative sample of 16 countries, differentiated by colors.}
		\label{real_co2}
    \end{center}
\end{figure*}

To illustrate this ambition, we provide on \Cref{real_co2} a visualisation of the raw observations for the first 16 countries appearing by alphabetical order in the data set (for the sake of clarity, we cannot display all countries at once).
While studying per capita quantities allows us to compare countries with different populations, we can observe that $CO_2$ emissions can massively differ around the world, and so does the regularity of measurements.
By simple inspection, some patterns seem to emerge for countries with similar topographical properties (oil and gas producing regions) or similar lifestyles (typically depending on the average wealth of populations). 
By applying \algoclust on this data set, we have noticed its ability to automatically recover patterns that seem \emph{logical} (although such a thing remains highly subjective). 
For instance, when setting $K = 5$, one cluster of 11 countries gathered United Kingdom, United States, Russia, France and Canada among others. 
Another cluster only counted Kuwait and Qatar as members, while the largest regrouped around 90 countries mainly from Africa, South Asia and South America, for which the $CO2$ emissions per capita have generally been really low during the past centuries.   
More generally, we reported the prediction performances on this dataset in \Cref{table_co2_data}. 
Notice that increasing the number of clusters dramatically improves accuracy in this context, as it seems clear on \Cref{real_co2} that some countries present atypical patterns and should be treated in separated clusters to reach satisfying predictions of the missing values.
Overall, these abilities to take advantage of group-structured data, even with irregular measurements, and to provide probabilistic statements, highlight once more the interest of \algoclust to tackle various machine learning problems both in supervised and unsupervised contexts. 

\begin{table}
\begin{center}
\begin{tabular}{c|cc|}
\cline{2-3}
 & Mean & $WCIC_{95}$ \\ \hline
\multicolumn{1}{|c|}{\algo} & 34.9 (89.5) & 92.9 (21.6) \\
\multicolumn{1}{|c|}{\algoclust - K = 2} & 28.9 (62.3) & 90.8 (24) \\
\multicolumn{1}{|c|}{\algoclust - K = 3} & 19.6 (49.1) & 92.5 (18.3) \\
\multicolumn{1}{|c|}{\algoclust - K = 4} & 15.4 (33.4) & 93.7 (17.8) \\
\multicolumn{1}{|c|}{\algoclust - K = 5} & \textbf{14 (28.7)} & \textbf{94.1 (17.1)} \\
\multicolumn{1}{|c|}{\algoclust - K = 6} & 14.2 (29.3) & 93.4 (18) \\ \hline
\end{tabular}
\end{center}
\caption{Average (sd) values of MSE and $WCIC_{95}$ for \algoclust with $K = 1, \dots, 6$ on the $CO_2$ emissions testing data set.}
\label{table_co2_data}
\end{table}

\section{Discussion}
\label{sec:conclusion}
We introduced a novel framework to handle clustering and regression purposes with a multi-task GPs mixture model.
This approach, called \algoclust, extends the previous algorithm \algo \citep{LeroyMAGMAInferencePrediction2020} to deal with group-structured data more efficiently.
The method provides new insights on the matter of GP regression by introducing cluster-specific modelling and predictions while remaining efficiently tractable through the use of variational approximations for inference.
Moreover, this nonparametric probabilistic framework accounts for uncertainty both in regards to the clustering and predictive aspects, which appears to be notable in the machine learning literature.
We demonstrated the practical efficiency of \algoclust on both synthetic and real data sets where it outperformed the alternatives, particularly in group-structured context.
Even though the main concern of our method remains the predictive abilities, the clustering performances also deserve to be highlighted, compared to state-of-the-art functional clustering algorithms.
\newline

While we recall that computational cost is of paramount importance to ensure broad applicability of GP models, the present version of \algoclust yet lacks a sparse approximation, which is not trivial to derive in this framework.
However, one of the state-of-the-art sparse methods \citep{TitsiasVariationalLearningInducing2009,BauerUnderstandingProbabilisticSparse2016} also makes use of variational inference, both to select pseudo-inputs and learn hyper-parameters of GP models. 
Therefore, an interesting extension could come from simultaneously computing $\mukk$'s hyper-posteriors and pseudo-inputs in \algoclust, allowing for a sparse approximation of the highest dimensional objects in our model. 
Besides, several additional features would be worth investigating in future work, such as the extension to non-Gaussian likelihoods or enabling online updates in the learning procedure.

\section{Proofs}
\label{sec:proofs}

As a prerequisite, let us introduce an intermediate result that will be used several times in the proofs below.

\begin{lemma}
\label[lemma]{lem:1}
	Let $X \in \mathbb{R}^N$ be a random Gaussian vector $X \sim \mathcal{N} \paren{ m , \mathbf{K} }$, where $\mathbb{E}_{X}$ denotes the expectation and $\mathbb{V}_{X}$ the variance with respect to this distribution. Additionally, let $b \in \mathbb{R}^N$ be an arbitrary vector and $\mathbf{S}$ a $N \times N$ covariance matrix. Then:
	
	\begin{align*}
		\mathbb{E}_{X } \croch{ (X - b)^{\intercal} \mathbf{S}^{-1} (X - b)}
		&= (m - b)^{\intercal} \mathbf{S}^{-1} (m - b) + \tr{\mathbf{K} \mathbf{S}^{-1}}.
	\end{align*}
\end{lemma}

\begin{proof}

	\begin{align*}
		\mathbb{E}_{X } \croch{ (X - b)^{\intercal} \mathbf{S}^{-1} (X - b)}
		&= \mathbb{E}_{X}\croch{\tr{\mathbf{S}^{-1}(X-b)(X-b)^{\intercal}}} \\
        &= \tr{\mathbf{S}^{-1} (m-b)(m-b)^{\intercal}} + \tr{\mathbf{S}^{-1}\mathbb{V}_{X}\croch{X}} \\
		&=  (m - b)^{\intercal}	\mathbf{S}^{-1}(m - b) + \tr{ \mathbf{K} \mathbf{S}^{-1} }.
	\end{align*}
\end{proof}

\subsection{Proof of \Cref{prop:E_step_Z}}
Throughout, we note $\mathbb{E}_{\mub}$ the expectation with respect to the variational distribution $\hat{q}_{\mub}(\mub)$.
From \citet[Chapter~10]{BishopPatternrecognitionmachine2006}, the optimal solution $\hat{q}_{\Z}(\Z)$ to the variational formulation verifies:
	\begin{align}
		\log \hat{q}_{\Z}(\Z) 
			&= \mathbb{E}_{\mub} \croch{ \log p(\yii, \Z, \mub \mid \hat{\Theta}) } + C_1 \nonumber \\
			&= \mathbb{E}_{\mub} \croch{ \log p(\yii \mid \Z, \mub , \thetaiihat, \sigmaiihat) + \log p(\Z \mid \piihat)  + \log p(\mub \mid \gammakkhat) } + C_1 \nonumber \\
			&= \mathbb{E}_{\mub} \croch{ \log p(\yii \mid \Z, \mub, \thetaiihat, \sigmaiihat) } + \log p(\Z \mid \piihat) + C_2 \nonumber \\ 
			&= \mathbb{E}_{\mub} \croch{ \sumi \sumk \Zik \log p(\yi \mid \Zik = 1, \mukti, \thetaihat, \sigmaihat) } +  \sumi \sumk \Zik \log( \pikhat) + C_2 \nonumber \\
			&=  \sumi \sumk \Zik \croch{ \log( \pikhat) + \mathbb{E}_{\mu_k} \croch{ \log p(\yi \mid \Zik = 1, \mukti, \thetaihat, \sigmaihat) } } + C_2 \nonumber \\
			&=  \sumi \sumk \Zik \Bigg[ \log( \pikhat) - \dfrac{1}{2}  \log \abs{\Psiihat^{\ti}} \nonumber \\ 
			& \hspace{2.5cm} - \dfrac{1}{2} \mathbb{E}_{\mu_k} \croch{ (\yi - \mukti)^{\intercal} \Psiihatti^{-1} (\yi - \mukti) }  \Bigg] + C_3. 	\label{eq_use_lemma}
	\end{align}
	
Applying \Cref{lem:1} to the expectation in \eqref{eq_use_lemma}, we obtain:

\begin{align*}
\log \hat{q}_{\Z}(\Z) 
	&= \sumi \sumk \Zik \Bigg[ \log( \pikhat) - \dfrac{1}{2} \paren{ \log \abs{\Psiihat^{\ti}} + \paren{\yi - \mkhat(\ti)}^{\intercal} \Psiihatti^{-1} \paren{\yi - \mkhat(\ti) } }\\
	& \hspace{2.5cm}  - \dfrac{1}{2} \tr{ \Chatkti \Psiihatti^{-1} }  \Bigg] + C_3 \\ 
	&= \sumi \sumk \Zik \croch{ \log \tauik }
\end{align*}
\noindent where (by inspection of both Gaussian and multinomial distributions): 
	\begin{equation*}
		\tauik = \dfrac{\pikhat \N \paren{\yi ; \mkhat(\ti) , \Psiihat^{\ti}} \exp\paren{ -\frac{1}{2} \tr{ {\Chatkti \Psiihat^{\ti}}^{-1} } } }{\sum\limits_{l = 1}^{K} \pilhat \N \paren{\yi ; \hat{m}_l(\ti) , \Psiihat^{\ti}} \exp\paren{ -\frac{1}{2} \tr{ { \Chatlti \Psiihat^{\ti}}^{-1}} }}, \ \forall i \in \I, \forall k \in \K.
	\end{equation*}

Therefore, the optimal solution may be written as a factorised form of multinomial distributions:
\begin{equation*}
	\hat{q}_{\Z}(\Z) = \prodi \mathcal{M} \paren{ \Zi; 1, \taui = (\uptau_{i1}, \dots, \uptau_{iK})^{\intercal} }.
\end{equation*}

\subsection{Proof of \Cref{prop:E_step_mu}}

Let $\mathbb{E}_{\Z}$ denote by the expectation with respect to the variational distribution $\hat{q}_{\Z}(\Z)$.
From \citet[Chapter~10]{BishopPatternrecognitionmachine2006}, the optimal solution $\hat{q}_{\mub}(\mub)$ to the variational formulation verifies:

\begin{align*}
	\log \hat{q}_{\mub}(\mub)
		&= \mathbb{E}_{\Z} \croch{ \log p(\yii, \Z, \mub \mid \hat{\Theta}) } + C_1 \\
		&= \mathbb{E}_{\Z} \croch{ \log p(\yii \mid \Z, \mub , \thetaiihat, \sigmaiihat) + \log p(\Z \mid \piihat)  + \log p(\mub \mid \gammakkhat) } + C_1 \\
		&= \mathbb{E}_{\Z} \croch{ \log p(\yii \mid \Z, \mub, \thetaiihat, \sigmaiihat) } + \log p(\mub \mid \gammakkhat) + C_2 \\
		&= \sumi \mathbb{E}_{\Zi} \croch{ \log	p(\yi \mid \Zi, \mub, \thetaihat, \sigmaihat) } +  \sumk \log p(\mukt \mid \gammakhat) + C_2 \\
		&= \sumi  \sumk \mathbb{E}_{\Zi} \croch{ \Zik } \log  p(\yi \mid \Zik = 1, \mukti, \thetaihat, \sigmaihat)  +  \sumk \log p(\mukt \mid \gammakhat) + C_2 \\
		&= - \dfrac{1}{2}\sumk  \Bigg[ (\mukt - \mkt)^{\intercal} \Ckhatt^{-1} (\mukt - \mkt)   \\
		& \hspace{1.8cm} + \sumi  \tauik  (\yi - \mukti)^{\intercal} \Psiihatti^{-1} (\yi - \mukti) \Bigg] + C_3. 
\end{align*}

If we regroup the scalar coefficient $\tauik$ with the covariance matrix $\Psiihatti^{-1}$, we simply recognise two quadratic terms of Gaussian likelihoods on the variables $\mu_k(\cdot)$, although evaluated onto different sets of timestamps $\Ut$ and $\ti$.
By taking some writing cautions and expanding the vector-matrix products entirely, it has been proved in \cite{LeroyMAGMAInferencePrediction2020} that this expression factorises with respect to $\mukt$ simply by expanding vectors $\yi$ and matrices $\Psiihatti$ with zeros, $\forall t \in \Ut, \ t\notin \ti$. 
Namely, we can notice that:
\begin{itemize}
	\item $\forall i \in \I, \ \tilde{\mathbf{y}}_i = \paren{\mathds{1}_{ [t \in \ti ]} \times y_i(t)}_{t \in \Ut}$, a $N$-dimensional vector,
	\item $\forall i \in \I, \ \Psitilde = \croch{ \mathds{1}_{ [t, t' \in \ti]} \times \psiihat \paren{t, t'} }_{t, t' \in \Ut}$, a $N \times N$ matrix.	\end{itemize}
Therefore: 

\begin{align*}
	\log \hat{q}_{\mub}(\mub) 
		&= - \dfrac{1}{2}\sumk \mukt^{\intercal} \paren{ \Ckhatt^{-1} + \sumi \tauik \Psitilde^{-1} } \mukt \\
		& \hspace{0.4cm} +  \mukt^{\intercal} \paren{ \Ckhatt^{-1} \mkt + \sumi \tauik \Psitilde^{-1} \tilde{\mathbf{y}}_i } + C_4.
\end{align*}

By inspection, we recognise a sum of a Gaussian log-likelihoods, which implies the underlying values of the constants. 
Finally:
\begin{equation}
	\hat{q}_{\mub}(\mub) = \prodk \N \paren{\mu_k(\Ut) ; \mkhat(\Ut) , \Chatkt},
\end{equation}
with: 
\begin{itemize}
	\item $\Chatkt = \paren{ \Ckhatt^{-1} + \sumi \tauik \Psitilde^{-1}}^{-1}, \ \forall k \in \K$, 
	\item $\mkhat(\Ut) = \Chatkt \paren{ \Ckhatt^{-1} \mkt + \sumi \tauik \Psitilde^{-1} \tilde{\mathbf{y}}_i },  \ \forall k \in \K$.
\end{itemize}

\subsection{Proof of \Cref{prop:M_step}}
Let $\mathbb{E}_{\Z, \mub}$ be the expectation with respect to the optimised variational distributions $\hat{q}_{\Z}(\Z)$ and $\hat{q}_{\mub}(\mub)$.
From \citet[Chapter~10]{BishopPatternrecognitionmachine2006}, we can figure out the optimal values for the hyper-parameters $\Theta$ by maximising the lower bound $\mathcal{L}(\hat{q};\Theta)$ with respect to $\Theta$:

\begin{equation*}
		\hat{\Theta} = \argmax\limits_{\Theta} \mathcal{L}(\hat{q}; \Theta).
\end{equation*}
Moreover, we can develop the formulation of the lower bound by expressing the integrals as an expectation, namely $\mathbb{E}_{\Z, \mub}$.
Recalling the complete-data likelihood analytical expression and focusing on quantities depending upon $\Theta$, we can write:

\begin{equation*}
\begingroup
\allowdisplaybreaks
\begin{aligned}
	\mathcal{L}(\hat{q}; \Theta)
		&= - \mathbb{E}_{ \{ \Z, \mub \} } \croch{  \underbrace{\log \hat{q}_{\Z, \mub}(\Z, \mub)}_{constant \ w.r.t. \ \Theta}  -\log p(\yii, \Z , \mub \vert \Theta)  } \\
		&= \mathbb{E}_{ \{ \Z, \mub \} } \croch{ \log  \prodk \acc{ \N \paren{\mukt;  \mkt , \Ckt }\prodi \paren{\pik \N \paren{ \yi; \mukti, \Psiiti }}^{\Zik} } } + C_1 \\
		&= \sumk \Bigg[ - \dfrac{1}{2} \paren{ \log \abs{\Ckt} + \mathbb{E}_{ \mub } \croch{ (\mukt- \mkt)^{\intercal} \Ckt^{-1} (\mukt- \mkt) } } \\
		& \hspace{1.3cm}  - \dfrac{1}{2} \sumi \mathbb{E}_{ \{ \Z, \mub \} } \croch{ \Zik \paren{ \log \abs{\Psiiti} +  (\yi - \mukti)^{\intercal} \Psiiti^{-1} (\yi - \mukti) } } \\ 
		& \hspace{1.3cm} + \sumi \mathbb{E}_{ \Z } \croch{ \Zik }   \log \pik   \Bigg] + C_2 \\
		&= \sumk \Bigg[ - \dfrac{1}{2} \paren{ \log \abs{\Ckt} + (\mkhat(\Ut)- \mkt)^{\intercal} \Ckt^{-1} (\mkhat(\Ut)- \mkt) + \tr{ \Chatkt \Ckt^{-1}} } \\
		& \hspace{1.3cm}  - \dfrac{1}{2} \sumi \tauik \paren{ \log \abs{\Psiiti} +  (\yi - \mkhat(\ti))^{\intercal} \Psiiti^{-1} (\yi - \mkhat(\ti)) + \tr{ \Chatkt \Psiiti^{-1}} } \\ 
		& \hspace{1.3cm} + \sumi  \tauik \log \pik  \Bigg] + C_2,
\end{aligned} 
\endgroup
\end{equation*}
	
\noindent where we made use of \Cref{lem:1} twice, at the first and second lines for the last equality.
By reorganising the terms on the second line, we can derive another formulation of this lower bound that allows for better managing of the computational resources. 
For information, we give this expression below since it is the quantity implemented in the current version of the \algoclust code: 

\begin{align*}
	\mathcal{L}(\hat{q}; \Theta)
	 	&= - \dfrac{1}{2} \sumk \Bigg[ \log \abs{\Ckt^{-1}} + (\mkhat(\Ut)- \mkt)^{\intercal} \Ckt^{-1} (\mkhat(\Ut)- \mkt) + \tr{ \Chatkt \Ckt^{-1}} \Bigg] \\
		& \hspace{0.4cm}  - \dfrac{1}{2} \sumi \Bigg[ \log \abs{\Psiiti} +  \yi^{\intercal} \Psiiti^{-1} \yi -2  \yi^{\intercal} \Psiiti^{-1} \sumk \tauik \mkhat(\ti) \\
		& \hspace{1.8cm}  +  \tr{ \Psiiti^{-1}  \sumk \tauik \paren{\mkhat(\ti) \mkhat(\ti)^{\intercal} +  \Chatkt } } \Bigg] \\ 
		& \hspace{0.4cm} + \sumk \sumi \tauik   \paren{ \log \pik }  + C_2.
\end{align*}

Regardless of the expression we choose for the following, we can notice that we expressed the lower bound $\mathcal{L}(q; \Theta)$ as a sum where the hyper-parameters $\gammakk$, $\{ \thetaii, \sigmaii \}$ and $\pii$ appear in separate terms.
Hence, the resulting maximisation procedures are independent of each other. 
First, we focus on the simplest term that concerns $\pii$, for which we have an analytical update equation.
Since there is a constraint on the sum $\sumk \pik = 1$, we first need to introduce a Lagrange multiplier in the expression to maximise:

\begin{equation}
\label{lagran}
	\lambda \paren{\sumk \pik - 1} + \mathcal{L}(q; \Theta).  
\end{equation}

Setting to 0 the gradient with respect to $\pik$ in \eqref{lagran}, we get:
\begin{align*}
	  \lambda + \dfrac{1}{\pik} \sumi \tauik = 0, \ \forall k \in \K.
\end{align*}

Multiplying by $\pik$ and summing over $k$, we deduce the value of $\lambda$: 
\begin{align*}
	  \sumk \pik \lambda &= - \sumk \sumi \tauik \\
	 \lambda \times 1  &= -  \sumi 1 \\
	 \lambda &= - M.
\end{align*}

Therefore, the optimal values for $\pik$ are expressed as:
\begin{equation}
	\pikhat  = \dfrac{1}{M} \sumi \tauik, \ \forall k \in \K.
\end{equation}
Concerning the remaining hyper-parameters, in the absence of analytical optima, we have no choice but to numerically maximise the corresponding terms in $\mathcal{L}(\hat{q}; \Theta)$, namely: 
\begin{equation}
\label{optimk}
- \dfrac{1}{2} \sumk \paren{ \log \abs{\Ckt^{-1}} + (\mkhat(\Ut)- \mkt)^{\intercal} \Ckt^{-1} (\mkhat(\Ut)- \mkt) + \tr{ \Chatkt \Ckt^{-1}} } ,
\end{equation}	
\noindent and 
\begin{equation}
\label{optimi}
- \dfrac{1}{2} \sumi  \sumk\tauik \paren{ \log \abs{\Psiiti} +  (\yi - \mkhat(\ti))^{\intercal} \Psiiti^{-1} (\yi - \mkhat(\ti)) + \tr{ \Chatkt \Psiiti^{-1}} }.
\end{equation} 
It is straightforward to see that some manipulations of linear algebra also allows the derivation of explicit gradients with respect to $\gammakk$, $ \thetaii$ and $\sigmaii$.
Hence, we may take advantage of efficient gradient-based methods to handle the optimisation process. 
We should stress that the quantity \eqref{optimk} is a sum on the sole values of $k$, whereas \eqref{optimi} also implies a sum on the values of $i$.
Hence, each term of these sums involves only one hyper-parameter at a time, which thus may be optimised apart from the others.
Conversely, if we assume all individuals (respectively all clusters) to share the same set of hyper-parameters, then the full sum has to be maximised upon at once.
Therefore, recalling that we introduced 4 different settings according to whether we consider common or specific hyper-parameters for both clusters and individuals, we shall notice the desired maximisation problems that are induced by \eqref{optimk} and \eqref{optimi}. 

\subsection{Proof of \Cref{prop:BIC}}
\label{sec:proof_BIC}

Let us reconsider the expression of $\mathcal{L}(\hat{q}; \Theta)$ from the previous proof. 
As the model selection procedure takes place after convergence of the learning step, we can use the optimal variational approximation $\hat{q}_{\Z, \mub}$ to compute the lower bound explicitly. 
Contrarily to the M step though, we now need to develop its full expression, and thus make use of \Cref{lem:1} three times.

\begingroup
\allowdisplaybreaks
\begin{align*}
	\mathcal{L}(\hat{q} ; \Theta)
		&= \mathbb{E}_{ \{ \Z, \mub \} } \croch{ \log p(\yii, \Z , \mub \mid \Theta) - \log \hat{q}_{\Z, \mub}(\Z, \mub) } \\
		&= \mathbb{E}_{ \{ \Z, \mub \} } \croch{ \log p(\yii \mid \Z , \mub, \thetaii, \sigmaii) } + \mathbb{E}_{ \Z } \croch{ \log  p \paren{\Z  \mid \pii) } } \\
		& \hspace{0.5cm} + \mathbb{E}_{ \mub } \croch{  \log  p \paren{\mub  \mid \gammakk) } } - \mathbb{E}_{ \Z } \croch{  \log \hat{q}_{\Z}(\Z) } - \mathbb{E}_{ \mub } \croch{ \log \hat{q}_{\mub}(\mub)}  \\
		&= \sumi \sumk  \acc{  \tauik  \paren{ \log \mathcal{N}\paren{ \yi; \mkhat(\ti)), \Psiiti } - \dfrac{1}{2}  \tr{  \Chatkt \Psiiti^{-1}} } }   \\
		& \hspace{0.5cm} + \sumi \sumk \acc{ \tauik \log \pik } + \sumk \acc{ \log \mathcal{N} \paren{ \mkhat(\Ut); \mkt , \Ckt }  - \dfrac{1}{2} \tr{ \Chatkt \Ckt^{-1}} }  \\
		& \hspace{0.5cm} - \sumi \sumk  \acc{ \tauik \log \tauik  } - \sumk  \acc{  \log \mathcal{N} \paren{ \mkhat(\Ut); \mkhat(\Ut) , \Ckt } - \dfrac{1}{2}  \tr{ \Chatkt \Chatkt {}^{-1}}  }    \\
		&= \sumi \sumk  \acc{  \tauik \paren{ \log \mathcal{N}\paren{ \yi; \mkhat(\ti), \Psiiti } - \dfrac{1}{2}  \tr{ \Chatkt \Psiiti^{-1}} } } \\
		& \hspace{0.5cm} + \sumk \acc{ \log \mathcal{N} \paren{ \mkhat(\Ut); \mkt , \Ckt }  - \dfrac{1}{2} \tr{ \Chatkt \Ckt^{-1}} + \dfrac{1}{2}   \log \abs{ \Chatkt } + N \log 2 \pi +  N  } \\
		& \hspace{0.5cm} + \sumi \sumk  \acc{ \tauik \log \dfrac{\pik}{\tauik}  }  \\
		&= \sumi \sumk  \acc{  \tauik \paren{ \log \mathcal{N}\paren{ \yi; \mkhat(\ti), \Psiiti } - \dfrac{1}{2}  \tr{ \Chatkt \Psiiti^{-1}} + \log \dfrac{\pik}{\tauik}  } } \\
		& \hspace{0.5cm} + \sumk \acc{ \log \mathcal{N} \paren{ \mkhat(\Ut); \mkt , \Ckt }  - \dfrac{1}{2} \tr{ \Chatkt \Ckt^{-1}} + \dfrac{1}{2}   \log \abs{ \Chatkt } + N \log 2 \pi +  N  }. \\
\end{align*}%
\endgroup

The result follows by considering the analogous expression $\mathcal{L}(\hat{q} ; \hat{\Theta})$ in which the hyper-parameters are evaluated at their optimal value. 


\subsection*{Acknowledgements} 
Significant parts of this work have been carried out while Arthur Leroy was affiliated with MAP5, Université de Paris, CNRS, UMR 8145, and Department of Computer Science, The University of Sheffield.
The authors warmly thank the French Swimming Federation for collecting data and sharing insights on the analysis, as well as Ai Ling Teh and Dennis Wang for providing data from the GUSTO project. The study is supported by the National Research Foundation (NRF) under the Open Fund-Large Collaborative Grant (OF-LCG; MOH-000504) administered by the Singapore Ministry of Health’s National Medical Research Council (NMRC) and the Agency for Science, Technology and Research (A*STAR). In RIE2025, GUSTO is supported by funding from the NRF’s Human Health and Potential (HHP) Domain, under the Human Potential Programme. Benjamin Guedj acknowledges partial support by the U.S. Army Research Laboratory and the U.S. Army Research Office, and by the U.K. Ministry of Defence and the U.K. Engineering and Physical Sciences Research Council (EPSRC) under grant number EP/R013616/1. Benjamin Guedj acknowledges partial support from the French National Agency for Research, grants ANR-18-CE40-0016-01 and ANR-18-CE23-0015-02.

\subsection*{Data availability}
The synthetic data, trained models and results are available at \url{https://github.com/ArthurLeroy/MAGMAclust/tree/master/Simulations}. The real data sets, associated trained models and results are available at \url{https://github.com/ArthurLeroy/MAGMAclust/tree/master/Real_Data_Study}. 

\subsection*{Code availability}
The current version of the R package implementing \algoclust is available on the CRAN and at \url{https://github.com/ArthurLeroy/MagmaClustR}.

\bibliography{biblio}   

\end{document}